\documentclass[10pt,journal,compsoc]{IEEEtran}
\ifCLASSOPTIONcompsoc
  \usepackage[nocompress]{cite}
\else
  \usepackage{cite}
\fi
\usepackage{subfigure}
\usepackage{diagbox}
\newtheorem{defn}{Definition}
 \usepackage{multirow}
 \usepackage{booktabs}
\usepackage{algorithmic}
\usepackage[ruled,linesnumbered]{algorithm2e}
\usepackage{graphicx}
\usepackage{balance}
\usepackage{url}
\usepackage{color}
\usepackage{threeparttable}
\usepackage{amssymb}
\usepackage{bbding}
\usepackage{pifont}
\usepackage{amsmath}
\usepackage{xcolor}

\ifCLASSINFOpdf
\else
\fi
\begin{document}
%
\title{Do as I can, not as I get:\\ Topology-aware multi-hop  reasoning \\ on  multi-modal knowledge graphs}
%
%
%
%

\author{Shangfei~Zheng,
        Hongzhi~Yin,
        Tong~Chen,
        Quoc Viet Hung~Nguyen,
        Wei~Chen,
        and~Lei~Zhao
\IEEEcompsocitemizethanks{

\IEEEcompsocthanksitem H. Yin and L. Zhao are the corresponding authors of this paper.


\IEEEcompsocthanksitem S. Zheng, W. Chen, and L. Zhao are with the Institute of Artificial Intelligence, School of Computer Science and Technology, Soochow University, Suzhou 215006, China.
E-mail: sfzhengsuda@stu.suda.edu.cn, \{robertchen, zhaol\}@suda.edu.cn.
\IEEEcompsocthanksitem H. Yin and T. Chen are  with the School of ITEE, The University of Queensland, St Lucia, QLD 4072, Australia. E-mail: db.hongzhi@gmail.com, tong.chen@uq.edu.au.
\IEEEcompsocthanksitem Q. Nguyen is with the School of Information and
Communication Technology,
Griffith University
Gold Coast, QLD 4215, Australia
E-mail: henry.nguyen@griffith.edu.au.

}

}

%
%

\markboth{Under Review}%
{Shell \MakeLowercase{\textit{et al.}}: Bare Demo of IEEEtran.cls for Computer Society Journals}

\IEEEtitleabstractindextext{%
\begin{abstract}
Multi-modal knowledge graph (MKG) includes triplets that consist of entities and relations and multi-modal auxiliary data.
In recent years, multi-hop multi-modal knowledge graph reasoning (MMKGR) based on reinforcement learning (RL) has received extensive attention because it addresses the intrinsic incompleteness of MKG in an interpretable manner.  However, its performance is limited by empirically designed rewards and sparse relations. In addition, this method has been designed for the transductive setting where test entities have been seen during training, and it works poorly in the inductive setting where test entities do not appear in the training set. To overcome these issues, we propose \textbf{TMR} (\textbf{T}opology-aware \textbf{M}ulti-hop \textbf{R}easoning), which can conduct MKG reasoning under  inductive and transductive settings. Specifically, TMR mainly consists of two components. (1) The topology-aware inductive representation captures information from the directed relations of unseen entities, and aggregates query-related topology features in an attentive manner to generate the fine-grained entity-independent features. (2) After completing multi-modal feature fusion, the relation-augment adaptive RL conducts multi-hop reasoning by eliminating manual rewards and dynamically adding actions. Finally,  we construct new MKG datasets with different scales for inductive reasoning evaluation. Experimental results demonstrate that TMP outperforms state-of-the-art MKGR methods under both inductive and transductive settings. 
\end{abstract}


}

\maketitle

\IEEEdisplaynontitleabstractindextext

%
\IEEEpeerreviewmaketitle

\IEEEraisesectionheading{\section{Introduction}\label{sec:introduction}}

%
%
%
%

\IEEEPARstart{K}{nowledge} graphs (KGs) store and manage huge amounts of data in reality and have been widely used in applications, including recommendation systems \cite{1:recom}, information retrieval \cite{2:IR}, and knowledge question answering \cite{3:QA}. A traditional KG consists of structural triplets that involve entities and relations, such as (\emph{James Cameron}, \emph{$Role\_create$}, \emph{$Rose Bukater$}). 
In recent years, as multi-modal data has received widespread attention in the field of data science and artificial intelligence, multi-modal knowledge graphs (MKGs) have emerged \cite{4:MMKG, 5:IKRL}. As shown in Figure 1(a), an MKG contains extra multi-modal auxiliary data (images and text description) based on structural triplets, which provides diverse modalities of knowledge. However, the intrinsic incompleteness of MKGs severely limits knowledge applications \cite{6:MTRL}.  To address this problem, the multi-modal knowledge graph reasoning (MKGR) technique is proposed to infer missing triplets in MKGs  \cite{7:MMKGR}. For instance, given a triple query (\emph{James Cameron}, \emph{Writer}, ?), MKGR can utilize both structural and multi-modal auxiliary data to infer the missing entity \emph{Titanic}.

\begin{figure*}
	\centering
	\subfigure[A small fragment of a training graph]
	{
		\centering
		\includegraphics[width=0.316\linewidth,height=4.25cm]{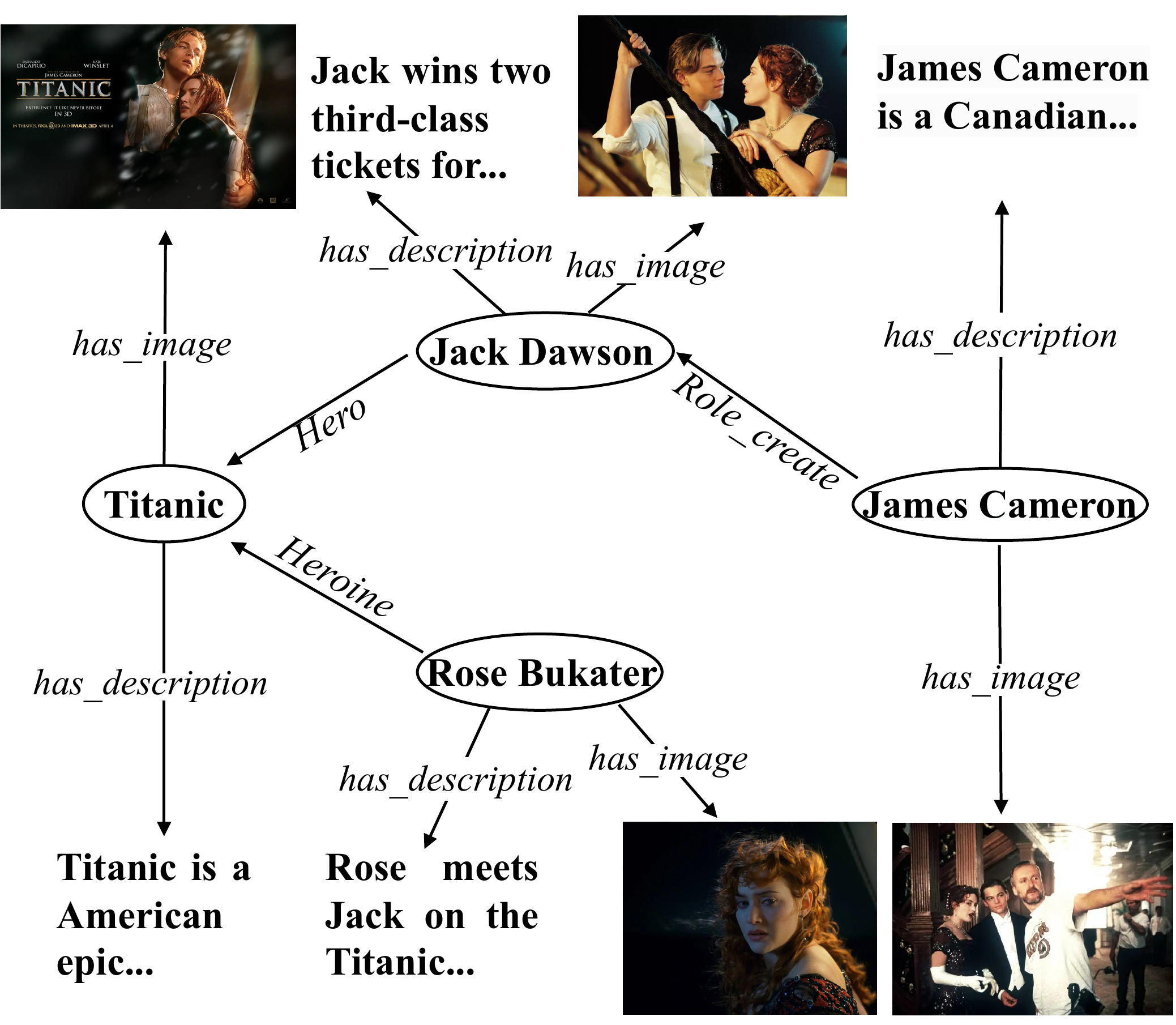}
	}
	\subfigure[MKGR under transductive setting]
	{
		\centering
		\includegraphics[width=0.316\linewidth,height=4.25cm]{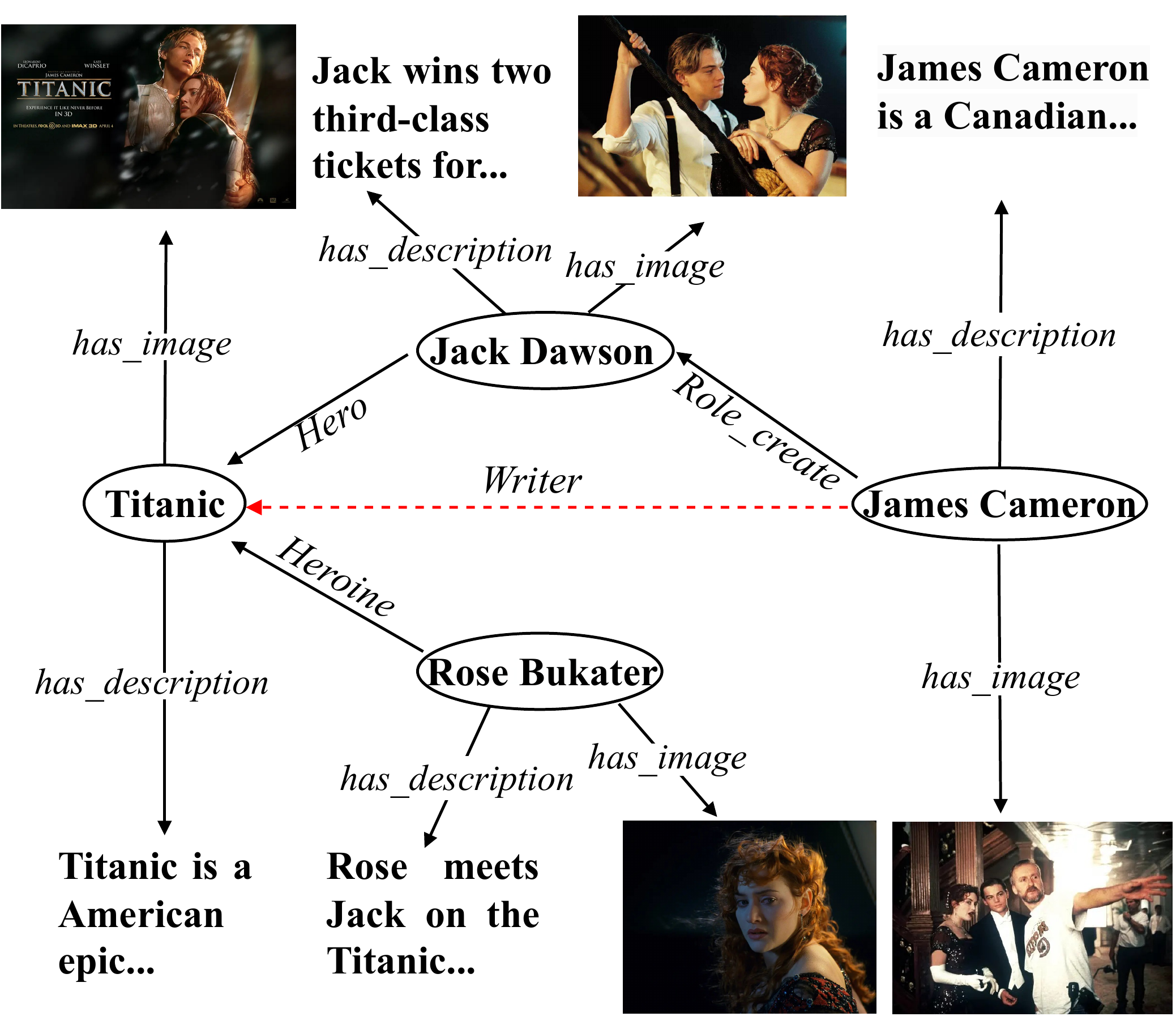}
	}
	\subfigure[MKGR under inductive setting]
	{
		\centering
		\includegraphics[width=0.32\linewidth,height=4.25cm]{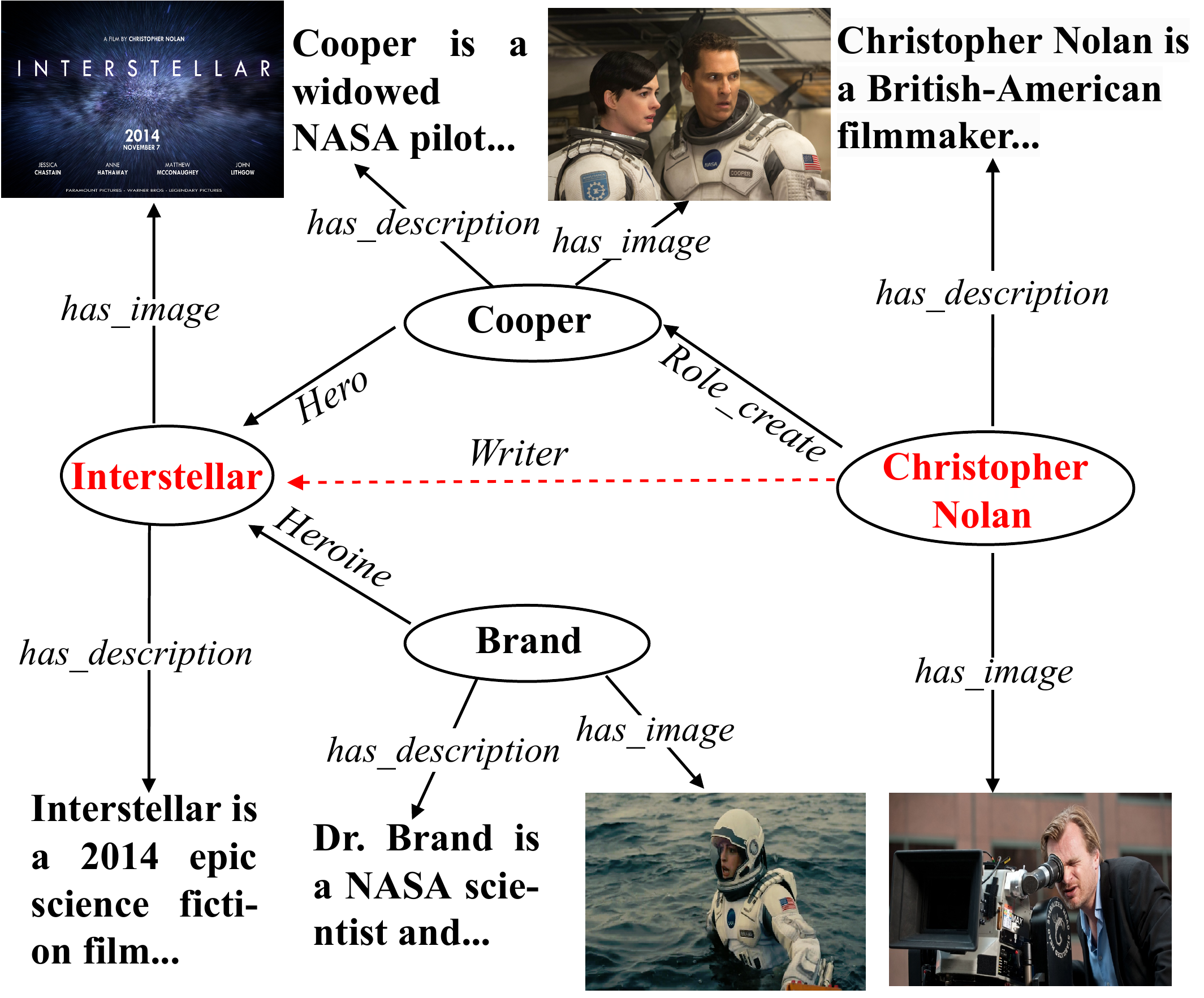}
	}
 \vspace{-0.2cm}
\caption{MKGR under transductive and inductive settings. In the transductive setting, test entities have been seen in the training graph. In contrast, testing entities such as \emph{Interstellar} and \emph{Christopher Nolan} did not appear in the training graph under the inductive setting. }
	\label{FIGURE1}
 \vspace{-0.4cm}
\end{figure*}

In the literature, existing MKGR methods can be categorized into two types: single-hop reasoning and multi-hop reasoning \cite{7:MMKGR}. The former focuses on modeling  score functions for one-step relations that contain relatively less information \cite{5:IKRL, 6:MTRL}, while the latter represents the latest work that interpretably infers missing elements by combining multi-hop relations and fusing the corresponding multi-modal features \cite{7:MMKGR}. As shown in Figure 1(b), by connecting (\emph{James Cameron}, \emph{Role\_{create}}, \emph{Jack Dawson}) and (\emph{Jack Dawson}, \emph{Hero}, \emph{Titanic}),  a missing triplet (\emph{James Cameron}, \emph{Writer}, \emph{Titanic}) can be inferred. 
MMKGR \cite{7:MMKGR} stands out as the unique multi-hop MKGR model in existing ones, garnering significant attention for its state-of-the-art (SOTA)   performance and interpretability. To effectively utilize both structural features and corresponding multi-modal features, MMKGR first uses a unified gate-attention network to generate multi-modal complementary features with sufficient attention interactions and less noise.  Then, these features are fed into a novel complementary feature-aware reinforcement learning (RL) framework. This framework selects a sequence of actions (i.e., multi-hop reasoning paths) to accumulate rewards on the basis of a manually designed 3D reward function. Finally, MMKGR aims to maximize reward values by successfully inferring missing entities and outputs interpretable reasoning paths. 

Although MMKGR demonstrates impressive reasoning performance and interpretability on MKGs, there is still scope for enhancing its action and reward design. (1) \emph{3D reward function is limited by manual design}. It comprises three manual  sub-rewards, relying on the experience of domain experts  and existing data distributions \cite{21:RARL}. However, this necessitates time-consuming redesign when adapting to new environments \cite{23:IRL}.  Moreover, the subjective nature of manual reward design can lead to variations among different designers \cite{22:RLR}. (2) \emph{MMKGR is  sensitive to the sparsity of relations}. The selection of actions in MMKGR relies on the combination of multi-hop relations. 
The absence of any relation in this path causes the reasoning path to be unavailable, which limits the reasoning performance  \cite{25:sparse1, 26:sparse2}. For example, MMKGR infers the missing entity \emph{Titanic} through a two-hop reasoning path \emph{James Cameron} $\stackrel{Role\_{create}}{\longrightarrow}$ \emph{Jack Dawson} $\stackrel{Hero}{\longrightarrow}$ \emph{Titanic}. If \emph{Role\_create} or \emph{Hero} is unconnected, the aforementioned two-hop path does not exist, which results in the query  (\emph{James Cameron}, \emph{Director}, ?) cannot be inferred. Arguably, it is extremely challenging to design an adaptive reward without manual intervention and dynamically add actions to alleviate sparsity.

More importantly, MMKGR is difficult to apply to real scenarios since it primarily concentrates on the transductive setting while overlooking the importance of the inductive setting. As shown in Figure 1 (b), all entities are assumed to be seen during testing in the transductive setting  \cite{10:GraIL}. However, knowledge is evolving and new entities are constantly emerging in the real world \cite{11:CoMPILE}. This observation is more in line with the inductive setting where the inferred entities in the test set do not appear in the training set \cite{12:LAN}. These inferred entities, often referred to as unseen entities, lack knowable structural representations under the inductive setting. Intuitively, MMKGR can leverage multi-modal auxiliary data of unseen entities to infer the missing triples associated with them.  This naturally raises an intriguing and fundamental question: How does MMKGR perform under the inductive setting?  To answer this question, we first construct datasets of the induction setting where the entities in the test set and the train set are disjoint \cite{13:DRUM}. Then, we apply MMKGR to conduct multi-hop reasoning under the inductive setting. Experimental results reveal that MMKGR struggles to converge and has low reasoning performance under the inductive setting. Actually, the conclusion of inductive reasoning methods \cite{14:SNRI, 13:DRUM} on traditional KGs is consistent with the above experimental findings: a transductive reasoning method that relies only on  multi-modal auxiliary  data and lacks generalizability to unseen entities is unsuitable  for the inductive setting \cite{12:LAN, 16:Indigo, 17:VNNet}.  This prompts a subsequent goal: \emph{How to develop the inductive capability of MMKGR to generalize unseen entities under an inductive setting?}

A technical challenge to achieving this goal lies in the lack of fine-grained entity-independent representations in existing MKGR methods. One of the key advantages of learning this representation is the development of the inductive capability to generalize unseen entities even in the absence of their specific structural features. \cite{10:GraIL, 16:Indigo}.  MMKGR lacking the inductive capability has no choice but to use multi-modal auxiliary features of unseen entities to understand these entities, which is highly dependent on the quality and quantity of multi-modal data and not suitable for unseen tasks \cite{12:LAN, 17:VNNet}. Additionally, existing entity-independent representation methods of inductive reasoning on traditional knowledge graph reasoning cannot be directly extended in MMKGR.  This is because these methods struggle to aggregate  the most relevant information based on specific query relations, resulting in the generation of coarse-grained representations for unseen entities. To make matters worse, the coarse-grained representation of each entity in the reasoning path iteratively disrupts decision-making abilities, which impairs the reasoning performance of MMKGR.  Consequently, the fine-grained entity-independent representation is crucial to develop an inductive capability for MMKGR.

In light of the aforementioned challenges in MMKGR, we propose an extended method entitled \textbf{TRM} (\textbf{T}opology-aware \textbf{M}ulti-hop \textbf{R}easoning).  The main difference between our method and existing ones is that TRM has not only a talent for exploiting multi-modal data, but also the inductive capability  to generalize unseen entities. Thus, TRM is capable of conducting MKGR under both inductive and transductive settings. 
Specifically, TRM mainly contains \emph{topology-aware inductive representation} (TAIR) and \emph{relation-augment adaptive reinforcement learning} (RARL). To develop the inductive capability for MMKGR, \textbf{TAIR} learns fine-grained entity-independent representation from query-related topology knowledge. Its relation-aware entity initializer   captures a coarse-grained entity-independent representation by leveraging type information of unseen entities from the connected directed relations. To further generate the fine-grained representation, an adaptive topology representation module introduces a query-aware graph neural network (GNN) to attentively capture the topological information. After completing multi-modal feature fusion, \textbf{RARL}  infers the missing elements by multi-hop reasoning path on MKGs, 
 aiming to further improve the complementary feature-aware reinforcement learning framework of MMKGR.
Technically, RARL not only dynamically adds relations as additional actions to eliminate relational sparsity but also adaptively generates rewards by imitating expert demonstrations with filtering low-contributing paths.  In summary, as an extension of our conference paper \cite{7:MMKGR}, this work makes the following contributions:

\begin{figure*}
 \centering 
  \label{Figure 2}
  \includegraphics[width=0.99\linewidth,height=6.8cm]{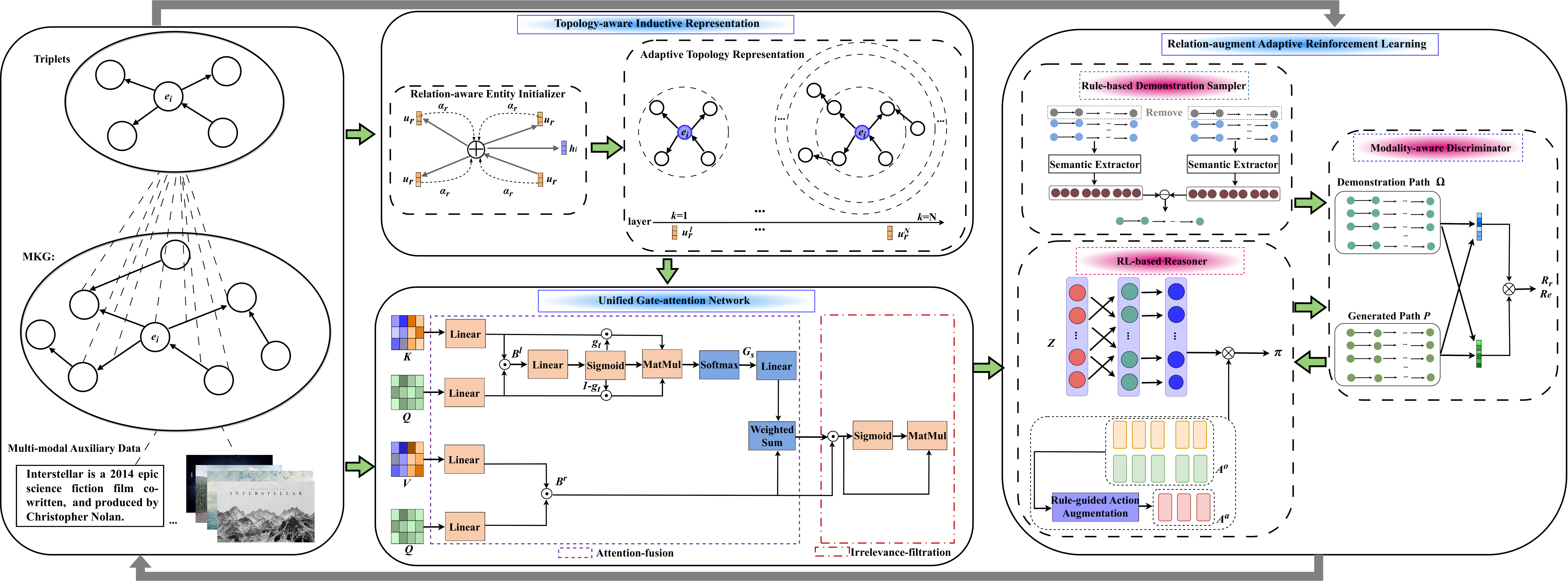}
 \hfill 
 \caption{TAIR first exploits query-related topological information to obtain fine-grained entity-independent  features. Then, these features and multi-modal auxiliary features are fed into the UGAN to generate multi-modal complementary features $Z$. Next, the RL-based reasoner utilizes $Z$ and the augmented actions to generate reasoning paths. The discriminator compares reasoning paths and demonstrations to output adaptive rewards for the reasoner. Finally, the  reasoner updates the reasoning policies and interacts with MKG to complete the prediction.}
\vspace{-7pt}
\end{figure*}

\begin{itemize}
\item 
To the best of our knowledge, this is the first work to investigate \emph{how to conduct MKGR under both  inductive and transductive settings.}

\item To resolve the above problem, we propose an RL-based MKGR model called TMR that mainly contains two components TAIR and RARL. Specifically, TAIR generates fine-grained entity-independent representations to generalize unseen entities. RARL conducts multi-hop reasoning by expanding action space and utilizing imitation learning to eliminate manually designed rewards.
\item  We construct MKG datasets under the inductive setting. To simulate  unseen entities, we ensure that the entities in the test set and training set are disjoint.
\item Extensive experiments are conducted under both the transductive and inductive settings. Experimental results demonstrate the superior performance of TMR surpasses MMKGR and various baselines.
\end{itemize}

The remaining sections are organized as follows. Preliminaries and definitions are presented in Section 2, followed by the overview of TMR. For different components in our proposed model, we introduce them in Sections 4, 5, and 6, respectively. Extensive experiments are shown in Section 7. Section 8 provides a review of the related literature. Finally, we conclude this work in Section 9.

\section{Preliminaries and Definitions}

A MKG is an extension of KG by adding multi-modal auxiliary data, it is  denoted as 
	\begin{math}
		\mathcal{G}_m = \{\mathcal{E}_m, \mathcal{R}, \mathcal{U}\} 
	\end{math}, where $\mathcal{R}$ is a set of semantic relations, and $\mathcal{E}_m$ denotes a set of entities   associated with related multi-modal auxiliary data. The features of an entity $i$ are denoted $\textbf{\emph{f}}_{i}$ = $\textbf{\emph{f}}_{s}$ $\circ$ $\textbf{\emph{f}}_{m}$, where  ``$\circ$" represents a multi-modal fusion method, $\textbf{\emph{f}}_{s}$ and $\textbf{\emph{f}}_{m}$ denote structural features and multi-modal auxiliary features, respectively. \begin{math}
		\mathcal{U} = \{ (\emph{$e_s$}, \emph{r}, \emph{$e_d$}) \mid \emph{$e_s$}, \emph{$e_d$} \in \mathcal{E}_m, \emph{r} \in \mathcal{R} \} 
	\end{math} is a set of triplets, where \emph{$e_s$}, \emph{$e_d$}, and \emph{$r$} denote a head entity, a tail entity, and the relation between these entities, respectively.  MKGR typically refers to the link prediction task of the inferring triple query ($e_s$, \emph{$r_q$}, ?) and (?, $r_{q}$, $e_d$), where \emph{$r_q$} is a query relation. By adding inverse relation, each triplet ($e_s$, \emph{r}, \emph{$e_d$}) is equivalent to the triplet (\emph{$e_d$}, $r^{-1}$, $e_s$).  Without loss of generality,  MKGR methods can predict  missing head entities by converting (?, \emph{$r_q$}, \emph{$e_d$}) to ($e_s$, \emph{$r_q^{-1}$}, ?).



 \begin{defn}
		\label{def:activity}
			\emph{MKGR under the  transductive setting}. 
Given a MKG $\mathcal{G}_{m}$ = $\{\mathcal{E}_m, \mathcal{R}, \mathcal{U}\}$, MKGR under transductive setting aims to reason out a set of triple queries  $\mathcal{Q}$, $\mathcal{Q}$ = \{(\emph{$e_s$}, \emph{$r_q$}, ?) $\mid$ (\emph{$e_s$}, \emph{$r_q$}, ?) $\notin$ $\mathcal{U}$, \emph{$e_s$}, ``?" $\in$ $\mathcal{E}_m$, \emph{$r_q$} $\in$ $\mathcal{R}$\}, where ``?" is a missing entity, $\mathcal{E}_m$ $\in$ $\mathcal{G}_{m}$ and $\mathcal{R}$ $\in$ $\mathcal{G}_{m}$ represent entity and relation have been seen in the existing MKG $\mathcal{G}_{m}$.
	\end{defn}

 \begin{defn}
		\label{def:activity}
			\emph{MKGR under the inductive setting}. 
Given two disconnected MKGs $\mathcal{G}_{m}$ = $\{$$\mathcal{E}_m$, $\mathcal{R}$, $\mathcal{U}$$\}$  and $\mathcal{G}_{m}^*$ = $\{$$\mathcal{G}_{m}^{*}$ = $\{\mathcal{E}_m^*, \mathcal{R}^{*}, \mathcal{U}^*\}$, $\mathcal{G}_{m}$ is often known as a training graph, while $\mathcal{G}_{m}^*$ is considered as a testing graph composed of the triplets by the emerging entities $\mathcal{E}_m^*$ and the relations $\mathcal{R}^{*}$.  MKGR under the inductive setting requires the model to learn  inductive capability in the training graph to infer a set of  queries $\mathcal{Q}$ on the test graph, $\mathcal{Q}$ = \{(\emph{$e_s$}, \emph{$r_q$}, ?) $\mid$ (\emph{$e_s$}, \emph{$r_q$}, ?) $\notin$ $\{$$\mathcal{U}$ $\cup$ $\mathcal{U}^*$$\}$, \emph{$e_s$}, ``?" $\in$ $\mathcal{E}_m^*$, \emph{$r_q$} $\in$ $\mathcal{R}^{*}$\}, where  $\mathcal{E}_m$ $\cap$ $\mathcal{E}_m^*$ = $\emptyset$ and $\mathcal{R}$ $\cup$ $\mathcal{R}^*$ = $\mathcal{R}$.
	\end{defn}

	\begin{defn}
		\label{def:activity}
			\emph{Multi-hop reasoning}. Multi-hop reasoning infers the missing element by the relational path shorter or equal $L$ hops, where $L$ is an integer not less than 1. A reasoning path is denoted  as $P$, which is obtained by summing all relations and entities in this path.
	\end{defn}

\section{Overview of TRM}
MMKGR, the version of our conference \cite{7:MMKGR}, is limited by manually designed reward functions and relation sparsity as well as poor performance under inductive settings, which motivates us to propose TMR in this paper.  As shown in Figure 2, TMR mainly contains two components: \textbf{TAIR} and \textbf{RARL}.  Specifically, inspired by the human inductive ability to generalize unseen tasks from existing relevant knowledge \cite{{27:ICHuman}}, \textbf{TAIR} generates fine-grained entity-independent features from the existing topological structure in an attentive manner to represent unseen entities.  After employing the unified gate-attention network in MMKGR to complete multi-modal feature fusion, \textbf{RARL} conducts MKGR by dynamically adding actions and automatically generating rewards, which is mainly inspired by the fact that humans learn optimal policies by imitating demonstrations rather than predefined paradigms \cite{28:IIhuman}. Notably, TMR is qualified for conducting MKGR under inductive and transductive settings. This is because TMR decouples representation and reasoning into independent components to ensure the flexibility of reasoning under different settings. When inductive settings are converted to transductive setting, TMR only needs to add additional structural representations of seen entities into the unified gate-attention network, while the reasoning module RARL continues to complete reasoning as a multi-modal perception interface without further changes.

 \section{Topology-aware Inductive Representation}
Existing methods are powerless to capture fine-grained entity-independent representation, thereby restricting the  inductive capability of MKGR models \cite{12:LAN, 16:Indigo, 17:VNNet}. To address the problem, we propose a novel representation method called TAIR in this section. Notably, the technical difference from the existing method for representing unseen entities lies in the following two points: (1) Taking full advantage of type information 
derived from the connected directed relations of unseen entities.
(2) Aggregating query-related neighbor relations in an attentive manner. Specifically, TAIR includes two modules, i.e., a relation-aware entity initializer and an adaptive topology representation. The former obtains coarse-grained representations of the unseen entity, and the latter aggregates topology information related to the query to generate ine-grained entity-independent representations.

 \subsection{Relation-aware Entity Initializer}

In general,  entities with similar semantics have similar topological structures in  MKGs, which are reflected in the connection patterns of their incoming and outgoing relations. By analyzing the connection patterns, we can obtain a coarse-grained representation that includes type information for unseen  entities. For example, unseen  entities $James\_Cameron$ and $Christopher\_Nolan$ both contain the outgoing edge $Role\_create$, and these two entities have the same type-level representation, i.e., $art\ creator$.
 
For an unseen entity $e_i$, its initialized embedding $\textbf{h}_i^{0}$ $\in$ $\mathbb{R}^{d}$ as follows:
\begin{equation}
 \textbf{h}_i^{0} = \frac{  \sum_{r \in I(i)} \textbf{W}_{i} \textbf{u}_r + \sum_{r \in O(i)} \textbf{W}_{o} \textbf{u}_r}{\left| I(i) \right| + \left| O(i) \right|}
\end{equation}
where $\textbf{W}_{i}$, $\textbf{W}_{o}$ $\in$ $\mathbb{R}^{d \times d}$ are transformation matrices. $I(i)$ and $O(i)$ represent the set of incoming and outgoing relations of the entity $e_i$, respectively. $\textbf{u}_r$ $\in$ $\mathbb{R}^{d}$ is embedding of the relation $r$. 

Considering that the semantics of the same entity can be diverse under different query relations \cite{29:DRDE}, we utilize the attention mechanism to filter out irrelevant neighbor relations. For example, the relations connected by  unseen entity \emph{Stephen Curry} have different types, such as family and vocational relations. Given a triple query (\emph{Stephen Curry}, \emph{Father}, ?), the vocational relations connected with the unseen entity indicate that \emph{Stephen Curry} is a professional basketball player, but this information is irrelevant to the query. Therefore, an attention mechanism that dynamically adjusts weights is employed to more accurately represent unseen entities in query tasks. The calculation process is as follows. 
\begin{equation}
 \alpha_r = softmax(\textbf{u}_r, \textbf{u}_{r_q}) = \frac{\rm{exp}(\textbf{u}_r^{\top}\textbf{u}_{r_q})}{\sum_{r^{'} \in \mathcal{N}^{r}(i)} \rm{exp}(\textbf{u}_{r'}^{\top}\textbf{u}_{r_q})}
\end{equation}
where $\textbf{u}_r$ and $\textbf{u}_{r_q}$ are  relation representations of neighbor relation $r$ and query relation $r_q$. $\alpha_r$ denotes the correlation between $r$ and $r_q$.

After integrating $\alpha_r$, Eq. (1) is updated as,  
\begin{equation}
 \textbf{h}_i^{0} = \frac{  \sum_{r \in I(i)} \alpha_r \textbf{W}_{i} \textbf{u}_r + \sum_{r \in O(i)} \alpha_r \textbf{W}_{o} \textbf{u}_r}{\left| I(i) \right| + \left| O(i) \right|}
\end{equation}

 \subsection{Adaptive Topology Representation}
After obtaining the coarse-grained representation $\textbf{h}^0$ of the unseen entity $e_i$ by the  initializer, we further capture the fine-grained semantic information from the topology of unseen entities.  Inspired by the ability of GNNs to capture topology information in knowledge graphs \cite{30:REDGNN}, an adaptive topology representation module  leverages  GNNs to aggregate local structural information from multi-hop neighbors of entity $e_i$. Specifically, we first concatenate the 
entities and their relations to obtain triplet information. Compared with individual entities or relations, triple information can provide sufficient topological information \cite{30:REDGNN}. Then, we compute the correlation between the query relation $r_q$ and these triplets that contain more contextual information, which effectively captures the fine-grained representation between topology and $r_q$ \cite{31:Fine1, 32:Fine2}. 
Next, we define updating process of the unseen entity $e_i$ in a \emph{k}-th layer as follows.
\begin{equation}
\begin{split}
 \textbf{h}_i^{k} = tanh(\textbf{W}_{self}^{k-1}\textbf{h}_i^{k-1} + \sum_{(i^{'}, r) \in \mathcal{N}_{i}(e_i)} \alpha_{i,r} \textbf{W}_{in}^{k-1} (\textbf{h}_i^{k-1} \circ \textbf{u}_r^{k-1}) \\ + \sum_{(r,i^{'}) \in \mathcal{N}_{o}(e_i)} \alpha_{i,r} \textbf{W}_{out}^{k-1}  (\textbf{h}_i^{k-1} \circ \textbf{u}_r^{k-1})) 
 \end{split}
\end{equation} 
\begin{equation}
 \alpha_{i,r} = \sigma (\textbf{W}_2 \textbf{c}_{i,r} + \textbf{b}) 
\end{equation}
\begin{equation}
 \textbf{c}_{i,r} = \sigma (\textbf{W}_1 [ \textbf{h}_i^{k-1} \oplus \textbf{h}_j^{k-1} \oplus \textbf{u}_r^{k-1} \oplus \textbf{u}_{r_q}^{k-1}])
\end{equation}
where $\mathcal{N}_{i}$ and $\mathcal{N}_{o}$ are the incoming and outgoing neighbors of entity $e_i$, respectively. $\textbf{W}_{self}^{k-1}$, $\textbf{W}_{in}^{k-1}$ and $\textbf{W}_{out}^{k-1}$ $\in$ $\mathbb{R}^{d \times d}$ denote the transformation matrices, respectively. $\circ$ is the element-wise product, and $\sigma$ is the activation function $sigmoid$. $\alpha_{i,r}$ is the attention weight of the triplet ($e_i$, $r$, $e_j$). Based on this, we obtain the fine-grained entity-independent representation $\textbf{h}_i^{k}$ of the unseen entity $e_i$.


Finally, to maintain the consistency of entities and relations within the embedding space, the embeddings of these relations are updated as follows:
\begin{equation}
 \textbf{u}_r^{k} = \textbf{W}_r^{k-1} \textbf{u}_r^{k-1}
\end{equation}
where $\textbf{W}_r$ $\in$ $\mathbb{R}^{d}$ is a transformation matrix.

\section{Unified Gate-attention Network}
In this section, we employ the unified gate-attention network (UGAN) in MMKGR to conduct feature fusion of fine-grained entity-independent
representation and multi-modal auxiliary representation. Specifically, the unified gate-attention network includes an attention-fusion module, and an irrelevance-filtration module.  After extracting multi-modal auxiliary features, the attention-fusion module fuses these features and context features together, by attending them with a carefully designed fine-grained attention scheme. Then, the irrelevance-filtration module discards irrelevant or even misleading information and generates noise robust multi-modal complementary features. Based on this, the unified gate-attention network selects features of different modalities online and simultaneously completes intra-modal and inter-modal attention interactions with noise robustness.

\subsection{Feature Extraction}
(1) Context features: The entity $e_l$ at reasoning step $l$ and query relation $r_q$ are represented as the fine-grained entity-independent embedding $\boldsymbol{h}_{l}$ and $\textbf{u}_{r_q}$, respectively. In addition, the history of the  reasoning path that consists of the visited entities and relations is defined as $b_l$ = ($e_s$, $r_0$, $e_1$, $r_1$,...,$e_l$). We leverage LSTM to integrate the vector of history information $\textbf{h}_l$  with \emph{$d_s$} dimensions into context features. Given the query in our multi-hop reasoning process, we obtain the context features  \textbf{y} = $[\textbf{b}_{l};\textbf{h}_l;\textbf{u}_{r_q}]$  by concatenating these features. Following \cite{7:MMKGR},  a group of context features $Y$ is calculated as 
\begin{math}
	Y = [\textbf{y}_1,\textbf{y}_2,...,\textbf{y}_{m}]
\end{math}, where \begin{math}
	Y\in\mathbb{R}^{m \times {d_y}}
\end{math}, \emph{m} and \emph{$d_{y}$} are the number of entities and the dimension of the features, respectively. (2) Multi-modal auxiliary features: To initialize image features $\textbf{f}_{i}$, we extract a \emph{$d_i$}-dimensional vector of the last fully-connected layer before the softmax in VGG model \cite{33:VGG}. Textual features $\textbf{f}_{t}$ are initialized by the word2vec framework \cite {34:word2Vec} and expressed as a \emph{$d_t$}-dimensional vector. We concatenate the above two parts of features on rows to form the multi-modal auxiliary features \begin{math}
	\textbf{x} = [\textbf{f}_{t}W_{t};\textbf{f}_{i}W_{i}]
\end{math}.  To flexibly add multi-modal auxiliary features, a group of \textbf{x} is denoted as  \begin{math}
	X = [\textbf{x}_{1},\textbf{x}_{2},...,\textbf{x}_{m}]
\end{math}, where $\emph{$W_t$} \in \mathbb{R}^{d_t \times {d_x}/2}$, $\emph{$W_i$} \in \mathbb{R}^{d_i \times {d_x}/2}$, and \begin{math}
	X \in \mathbb{R}^{m \times {d_x}}
\end{math} represents a group of multi-modal auxiliary features,
\emph{$d_{x}$} is the dimension of the feature.

\subsection{Attention-fusion Module}
To obtain the complementary features with sufficient  interactions and less noise, we need to fuse the context features $Y$ and multi-modal auxiliary features $X$ generated in feature extraction. 
However, redundant features tend to have a negative impact on the prediction during the multi-modal fusion \cite{42:fusion}. 	Specifically, redundant features are either shifted versions of the features related to the triple query or very similar with little or no variations, which can amplify the negative effects of noise  \cite{43:fusion2}.  The redundant features add computational complexity and cause collinearity problems \cite{44:fusion3}. Consequently, we propose the attention-fusion module that fuses the context features and multi-modal auxiliary features effectively.

Specifically, we first utilize linear functions to generate the queries \emph{Q}, keys \emph{K}, and values \emph{V} of the attention mechanism,
\begin{equation}
	Q = XW_{q}, K= YW_{k}, V= YW_{v}
\end{equation}
where 
\begin{math}
	W_{q}\in\mathbb{R}^{d_x \times d}, W_{k}, W_{v}\in\mathbb{R}^{d_y \times d},\end{math}  and \begin{math} Q, K, V \in \mathbb{R}^{m \times d}
\end{math} have the same shape. Then, the joint representation $B^l$ of \emph{Q} and \emph{K} is learned based on the MLB pooling method \cite{61:mlb}, inspired by the recent successes of it in fine-grained multi-modal fusion,
\begin{equation}
      B^l = KW_{k}^{l} \odot QW_{q}^{l}
\end{equation} Similarly, we can generate the joint representation $B^r$ of \emph{V} and \emph{Q} with the following equation,
\begin{equation}
	B^r = VW_{v}^{r} \odot QW_{q}^{r}
\end{equation}
where $W_{k}^{l}, W_{q}^{l}, W_{v}^{r}, W_{q}^{r} \in \mathbb{R}^{d \times j}$ are embedding matrices, and $\odot$ is Hadamard product.

Next, the filtration gate $g_t$ applied to different feature vectors is defined as,
\begin{equation}
	 g_t = \sigma(B^lW_{m}) 
\end{equation}
where $W_{m} \in \mathbb{R}^{j \times d}$ is an embedding matrix and $\sigma$ denotes the sigmoid activation. Based on the filtration gate $g_t$, we can filter out the redundant features generated during fusion and obtain a new representation with the following probability distributions,
\begin{equation}
	G_s = softmax((g_t \odot K) ((1-g_t)\odot Q))
\end{equation}
where $g_t$ and $1-g_t$ are used to trade off how many context features and multi-modal auxiliary features are fused.

Finally, our attention-fusion module generates the attended features \emph{$\hat{V}$}=\{${\textbf{v}_i}\}_{i=1}^{m}$ by accumulating the enhanced bilinear values of context features and multi-modal auxiliary features,
\begin{equation}
	\emph{$\hat{V}$} = \sum \nolimits_{i=1}^m (G_sW_{g}^{l})B_i^r
\end{equation}
where $W_{g}^{l}\in \mathbb{R}^{d \times 1}$, and $\textbf{v}_i \in \mathbb{R}^{1 \times j}$ denotes a row of the attended  features $\emph{$\hat{V}$} \in \mathbb{R}^{m \times j}$, feature vector $B_{i}^{r}\in \mathbb{R}^{1 \times j}$ is a row of the embedding matrix $B^{r}$.

By designing the attention-fusion module, we can complete the intra-modal and inter-modal feature interactions in a unified manner at the same time. This is because the inputs of this module are pairs from context features and multi-modal auxiliary features, where each vector of a pair may be learned from the same modality or different ones.

\subsection{Irrelevance-filtration Module}
We use an irrelevance-filtration module to further improve the robustness of the model. The attended features $\hat{V}$ obtained by the attention-fusion module may contain irrelevant features \cite{45:fusion4}. Specifically, irrelevant features are irrelevant to the triple query in the reasoning process. Since the attention mechanism assigns weights to all features, these features tend to participate in model computation and mislead the reasoning policy \cite{46:fusion5}. 
This motivates our model to weight more on the most related complementary features and dynamically filter irrelevant ones. This is achieved by a well-designed irrelevance-filtration gate function. The output of this gate is a scalar, the value range of which is (0,1). The  multi-modal complementary features $\emph{Z}$ are obtained as follows,
\begin{equation}
	G_f = \sigma(B^r\odot\hat{V})
\end{equation}
\begin{equation}
	Z = G_f(B^r\odot\hat{V})
\end{equation}
where $\sigma$ and $G_f$ denote
the sigmoid activation function and  irrelevance-filtration gate, respectively.

 \section{Relation-augment Adaptive reinforcement learning}

The existing RL-based MKGR method is limited by manual rewards and sparse relations \cite{35:RLsurvey, 36:RLsurvey2}. To address this problem, we propose a novel RL-based framework entitled RARL in this section.   Compared with MMKGR, the main technical difference of RARL lies in the following two points. (1)  We effectively increase the additional actions to alleviate the negative impact of sparse relations on the RL-based model. (2) RARL utilizes generative adversarial imitating networks to adaptively learn rewards by imitating demonstrations, which can stabilize reasoning performance and eliminate manual intervention in reward design.  This provides a new research perspective for RL-based reasoning methods for MKGR.

RARL consists of three modules namely RL-based reasoner, rule-based demonstration sampler, and modality-aware discriminator.  Specifically,  the reasoner leverages a rule-guided action augmentation method that dynamically adds additional actions and outputs diverse generated paths about missing elements. Then, the rule-based demonstration sampler filters out low-contributing paths as well as extracts trustworthy demonstrations from MKGs. Next, the modality-aware discriminator  generates rewards to update the reasoner by evaluating the semantic similarity between demonstrations and reasoning paths. After sufficient adversarial training, the RL-based reasoner tries to deceive the discriminator to 
gain more adaptive reward values by imitating the demonstrations. We introduce the above three modules in subsections 6.1, 6.2, and 6.3, respectively.
 
\subsection{RL-based Reasoner}
\label{6.1}
\subsubsection{Reinforcement Learning Formulation}

RARL trains an agent to interact with the  with MKGs by modeling by Markov decision process (MDP). The MDP consists of a 4-tuple, i.e., States, Actions, Transition, Rewards. The agent selects actions based on the current state and obtains rewards from the environment (MKGs) to update its behavior policy until it reaches a termination state or a predefined reasoning step.

\textbf{States}:\quad The state of the agent at reasoning step \emph{l} is denoted as \emph{$s_l$}= (\emph{$e_l$}, (\emph{$e_s$}, \emph{$r_q$}) $\in$ $\mathcal{S}$, where $\mathcal{S}$ denotes a state space and \emph{$e_l$} represents the entity at the current 
 reasoning step \emph{l}. The source entity \emph{$e_s$} and the query relation \emph{$r_q$} are the global context shared throughout all steps.

\textbf{Actions}:\quad  For the given state \emph{$s_l$}, its original action space is the set of usable actions \emph{$A^{o}_l$} at reasoning step \emph{l} is 
expressed as \emph{$A^{o}_l$} = \{(\emph{$r_{l+1}$}, \emph{$e_{l+1}$})$\mid$ (\emph{$e_l$}, \emph{$r_{l+1}$}, \emph{$e_{l+1}$}) $\in$  $\mathcal{G}_m$\}. To alleviate relation sparsity, the rule-guided action augmentation module  adds extra potential actions \emph{$A^{a}_{l}$} into the original action space. Thus, the joint action space \emph{$A^{o}_l$} $\cup$ \emph{$A^{a}_l$} = $A_l$ $\in$ $\mathcal{A}$. In addition, we add 
\emph{STOP} action to avoid infinitely unrolling in the reasoning process. The \emph{STOP} action executes a self-loop  when the reasoning step is unrolled to the maximum step \emph{L}.

\textbf{Transition}:\quad  $\mathcal{P}_r$ is defined  to facilitate the transition from the current state \emph{$s_t$} to the next state \emph{$s_{l+1}$}. $\mathcal{P}_r$: $\mathcal{S}$ $\times$ $\mathcal{A}$ $\rightarrow$ $\mathcal{S}$ is defined as $\mathcal{P}_r$ (\emph{$s_l$}, \emph{$A_l$}) = $\mathcal{P}_r$ (\emph{$e_l$}, (\emph{$e_s$}, \emph{$r_q$}),  \emph{$A^o$}, \emph{$A^{a}$}). 

\textbf{Rewards}: Different from existing manually-designed reward functions, we design an adaptive reward mechanism to eliminate manual intervention, which achieves high reasoning performance in complex and uncertain environments. The adaptive reward arises from path comparisons between the generator and expert demonstration, and it is defined in Eq. 29.

\textbf{Policy Network} The policy function $\pi$ is used as a multi-modal perception interface to output the next action with the highest executable probability. For a given state, $\pi$ selects the promising action with the maximum likelihood, which is defined as, 
\begin{equation}
    \emph{$\pi$}_\theta(\emph{a}_l|\emph{s}_l) = softmax (\textbf{A}_{l}(\textbf{W}\text{ReLu}(Z)))
\end{equation}
where $\emph{a}_l$$\in$$A_l$, and $A_l$ can be encoded to $\textbf{A}_{t}$ by stacking the representations of existing available actions. 

\subsubsection{Rule-guided Action Augmentation}
\label{6.2}
The existing RL-based reasoning method assumes sufficient relation paths between entities, and regards these relations and connected tail entities as the next action. However, the intrinsic incompleteness of MKGs leads to sparse relations of an entity. Especially, emerging entities are sparsely connected to existing entities under the inductive setting. This sparsity limits the utilization of potential reasoning paths. Therefore, it is necessary to design an action space augmentation  method to eliminate the sparsity of the action space. Although the idea of augmenting the action space is promising, a major challenge is how to \emph{efficiently augment additional  actions}.

Intuitively, enumerating all relations and entities to compose an additional action space can complement the existing action space.  However, this combined search space is close to  $\mathcal{O}$($\lvert$$\mathcal{E}$$\rvert$$\times$$\lvert$$\mathcal{R}$$\rvert$), where $\lvert$$\mathcal{E}$$\rvert$ and $\lvert$$\mathcal{R}$$\rvert$ are the numbers of entities and relations in a MKG, respectively. For a large-scale MKG with millions of entities and thousands of relations, large search space becomes an intractable problem.  To address the above problem, we propose a novel action  augmentation method to efficiently augment additional actions. For a state $s_l$, the candidate set of augmented action is denoted as $C_t$ = $\{$$(r',e')|r'$ $\in$ $\mathcal{R}$ $\wedge$ $e'$ $\in$ $\mathcal{E}_m$ $\wedge$ $(e_l, r', e')$ $\notin$ $\mathcal{G}_{m}$$\}$. First, we calculate the probability for the candidate set $C_t$:
\begin{equation}
p((r',e')|s_l)= p(r'|s_l)
p(e'|r',s_l)
\end{equation}

To reduce the candidate action space and time-consuming, an approximate pruning strategy is used to filter out additional actions. The pruning strategy consists of  additional relation  selection using $p(r'|s_l)$ and entity generation using $p(e'|r',s_l)$.

Then, for the state $s_l$, the attention score of the candidate relations is calculated as $p(r|s_l)$,
\begin{equation}
\textbf{w} = softmax (MLP(\textbf{s}_l) \cdot [\textbf{u}_{r_1} ,...,\textbf{u}_{r_{\lvert \mathcal{R} \rvert}}])
\end{equation}
where \textbf{w} denotes the attention vector. We select top $x$ relations with the largest attention values in $\textbf{w}$ to obtain  additional relation set $\mathcal{R}_{add}$ = {$r^1$, $r^2$,..., $r^{I}$}. 

Next, we leverage AnyBURL \cite{53:AnyBURL} to mine logical rules and corresponding confidence scores from the MKGs to guide the generation of tail entities. A logical rule $\mathcal{B}$ is defined as:
 \begin{equation}
	\wedge_{i=1}^{N}r_i(e_i, e_{i+1}) \Rightarrow r_h(e_1, e_{N})  
\end{equation}
where $\wedge$ is logic conjunction, $N$ denotes the maximum length of the rule chain. The left side is a rule body, which is represented by a conjunction of atoms ($e_1$, $r_1$, $e_2$) $\wedge$ ... $\wedge$ ($e_i$, $r_i$, $e_{i+1}$) $\wedge$ ... $\wedge$ ($e_{N-1}$, $r_{N-1}$, $e_{L}$). In addition, the right-side is called a rule head, $r_h$ denotes the head relation. 

We further measure the quality of each $\mathcal{B}$ by a confidence score, it is calculated as:
 \begin{equation}
	conf(\mathcal{B}) =   \frac{pos(\mathcal{B})}{pos(\mathcal{B})+neg(\mathcal{B})}
\end{equation}
where $pos(\mathcal{B})$ and $neg(\mathcal{B})$ are the numbers of positive and negative trajectories, respectively.

Finally, we predict the tail entities $p(e'|r',s_l)$ by retrieving the mined rules. For an additional relation $r'$, we select the mined rules with the highest confidence $conf(\mathcal{B}$) to select the tail entity, which can complete  new facts. The fact set $\mathcal{F}$ can be formulated as, 
 \begin{equation}
 \begin{split}
	\mathcal{F} =  \{ (e_l, r', e')| \exists \{ (e_l, r_1, e_1), (e_1, r_2, e_2),..., (e_n, r_n, e') \\ \in \mathcal{K}  \wedge (r_1 \rightarrow r_2 \rightarrow ... \rightarrow r_n) \in \mathcal{B} \wedge r' \in \mathcal{R}\} \}
 \end{split}
\end{equation}
where $\mathcal{K}$ is a set of atoms. We add additional actions to the multi-hop reasoning process, which effectively  eliminates
the sparsity of the action space. With this implementation, our overall search space of action augmentation is $\mathcal{O}$($\lvert$$\mathcal{R}$$\rvert$+$\lvert$\emph{I}$\rvert$$\times$$\lvert$$\mathcal{K}$$\rvert$), where $\emph{I}$ and $\mathcal{K}$ $\ll$ $\mathcal{R}$.

\subsection{Rule-based Demonstration Sampler}
\label{6.3}


Inspired by the fact that learning reasoning rules can increase the trustworthiness of the demonstrations  \cite{40:KGSurvey}, we extract a set of rule paths $P_{r_q}$ of the query relation $r_q$ as candidate demonstrations. Then, we reserve a set of the shortest paths $P'_{r_q}$ with size $N$ to filtrate noisy paths. This is because the combination of the shortest relation path can accurately represent the semantic link between two entities \cite{37:shortp, 38:DeepPath}.  And longer paths are more likely to contain worthless
or even noisy information \cite{39:MultiHop}. Next, we encode the path package $\mu$ by concatenating all path   embeddings in $P'_{r_q}$. Inspired by ConvKB showcased the effectiveness of convolutional neural networks (CNNs) in extracting semantic features from paths, the path package denoted as  $\boldsymbol{\mu}$ $\in$ $\mathbb{R}^{Nd}$ is fed into a convolutional layer by sliding a kernel $\boldsymbol{\omega}$:
 \begin{equation}
 \label{sem}
	\boldsymbol{c}_1 =  W_{2} \rm Relu (W_{1} \rm ReLU(Conv(\boldsymbol{\mu}, \boldsymbol{\omega})))
\end{equation}

To further select the rule paths related to $r_q$, we design a counterfactual filtering method. We use the dot product to extract the correlation between $\textbf{u}_{r_q}$ and $\boldsymbol{c}_1$.
 \begin{equation}
 \label{simi}
	t_{\mu} = \sigma (\boldsymbol{c}_1 \cdot \textbf{u}_{r_q})
\end{equation}
where $t_{\mu}$ is the correlation score to evaluate the path package representation $\boldsymbol{c}_1$ and the query relation $\textbf{u}_{r_q}$. The counterfactual filtering method removes each rule path $x'$ from the  $P'_{r_q}$ in turn. The altered path set is calculated by Eq.~\ref{simi} to obtain a new correlation score. we refine the
demonstration set according to this score:
 \begin{equation}
	\Omega = \{x|t_{u}-t_{u-x'}>0, x\in x'\}
\end{equation}
where $\Omega$ is the refined demonstration containing $N'$ paths. If the path $x'$ is important to the query relation, the removal of $x'$ will lead to the result of $t_{u}-t_{u-x'}$ being positive.

\subsection{Modality-aware Discriminator}

Typically, relations in paths mainly focus on connections between entities, while entities focus on multi-modal representations. When evaluating demonstrations and generated paths, dividing entities and relations into two levels not only avoids information confusion but also better adapts to the variation of semantic features at different levels. 

\textbf{Relation Level} The discriminator aims to judge whether the demonstrations and generated paths are similar in relational connectivity. For a triple query, a set of reasoning paths $P$ are packaged by concatenating all path embeddings and denoted as $\nu$.

For a triple query, we first remove entities in $\Omega$ and $P$ to extract relation paths by summing all relations. By using the concatenation operation to package the path, we obtain relation packages $\nu_{\Omega}$ and $\nu_{P}$.  Next, we extract semantic features $\boldsymbol{c}_{\Omega}$ and $\boldsymbol{c}_{P}$  using Eq.~\ref{sem}. Finally, the sigmoid function is used to the features to the interval (0,1). 
 \begin{equation}
 \label{D}
	D(\boldsymbol{\nu_{\Omega}}) = \sigma (W_{s2} ReLU(W_{s1} \boldsymbol{c}_{\Omega}))
\end{equation}
where $\sigma$ is a sigmoid function. We can obtain the representation $D(\boldsymbol{\nu_{\Omega}})$ of $\boldsymbol{c}_1$ in the same calculation form with $D(\boldsymbol{\nu_{P}})$.

\textbf{Entity Level} To better identify the differences between multi-modal features of entities, we leverage multi-modal complementary features in paths to accomplish entity-level discrimination. Specifically, for a triple query, we first gather the entities in the set of  paths in $\Omega$ and $P$. Then, we also use the concatenation operation to package the entities to obtain two entity packages $\kappa_{\Omega}$ and $\kappa_{P}$. Finally, we employ a multi-layer neural network as the feature extractor, which is defined as follows,
 \begin{equation}
 \label{E}
	D(\boldsymbol{\kappa_{\Omega}}) = \sigma (W_{e} tanh (\boldsymbol{\kappa}_{\Omega}))
\end{equation}
where  $D(\boldsymbol{\kappa_{\Omega}})$ is formulated as the package features of entities. We can obtain $D(\boldsymbol{\kappa_{P}})$ n the same calculation form as  $D(\boldsymbol{\kappa_{\Omega}})$.

\subsection{Training}

In this section, we define the final reward function and formalize the training process. Specifically, we first introduce path features whose quality is higher than random noise into the reward. 
 \begin{equation}
	R_r = \max(D(\boldsymbol{\nu_{P}}) - D^{N}(\boldsymbol{\nu_{P}}), 0)
\end{equation}
 \begin{equation}
	R_e = \max(D(\boldsymbol{\kappa_{P}}) - D^{N}(\boldsymbol{\kappa_{P}}), 0)
\end{equation}
where $D^{N}(\boldsymbol{\nu_{P}})$ and $D^{N}(\boldsymbol{\kappa_{P}})$ respectively denote noise embeddings 
from the entity and relation layers. These embeddings consist of random noise sampled from a continuous uniform distribution. $R_r$ and $R_e$ represent the adaptive rewards from relation layer and entity layer, respectively.

Then, we define the adaptive reward $R(s_l)$ at the state  $s_l$ as follows:
 \begin{equation}
	R(s_l) = \alpha R_e + (1-\alpha) R_r
\end{equation}
where $\alpha$ is a balance factor. Note that, the agent  learns adaptive rewards from demonstrations without manually designing and tuning, which reduces manual intervention and subjective bias \cite{23:IRL}. In addition, the adaptive reward improves the generalizability of our proposed model and it is  suitable for unseen tasks. This is because the adaptive reward mechanism automatically capturing common meta-knowledge by learning relation patterns and multi-modal features in demonstrations \cite{41:Rewardada}.

Next, we optimize the modality-aware discriminator by reducing training loss and expect it to possess expertise in distinguishing between $P$ and $\Omega$. 
 \begin{equation}
	\mathcal{L}_r = D(\boldsymbol{\nu_{P}}) -D(\boldsymbol{\nu_{\Omega}})+ \lambda (\parallel \bigtriangledown_{\hat{\emph{p}}}D(\hat{\emph{p}})\parallel_{2} -1)^{2}
\end{equation}
where $\lambda$ is a penalty term and $\hat{\emph{p}}$ is sampled uniformly along straight lines between the generated path and the demonstration. 

For the discriminator at the entity level, we define loss $\mathcal{L}_r$  as follows:
 \begin{equation}
	\mathcal{L}_e = -(\log D(\boldsymbol{\kappa_{\Omega}}) +\log(1-D(\boldsymbol{\kappa_{P}})))
\end{equation}

Finally, to maximize the accumulated rewards of adaptive reinforcement learning and obtain the optimal policy, the objective function is as follows,
\begin{equation}
\mathcal{J}(\theta) = \mathbb{E}_{(e_s,r,e_d) \sim \mathcal{G}_{m}}\mathbb{E}_{a_1,...,a_{L}\sim {\emph{$\pi$}_\theta}}[\emph{R} (s_l\mid\emph{e}_s,r)]
\end{equation}

\section{Experiment}
\subsection{Datasets}

\begin{table}

	\centering
	\caption{Statistics of the experimental datasets over the transductive setting.}
 	\vspace{-0.2cm}
	\begin{tabular}{llllll}
		\hline
		Dataset     & \#Ent  & \#Rel        & \#Train &\#Valid &\#Test  \\
		\hline
		WN9-IMG-TXT       & 6,555  & 9     & 11,747& 1,337 & 1,319\\
		FB-IMG-TXT       & 11,757  & 1,231      & 285,850 & 29,580 & 34,863\\
		
		\hline
	\end{tabular}
	
	\label{tab:plain}
 \label{tabled1}
 \vspace{-0.5cm}
\end{table}	

\begin{table}
\tabcolsep 3.6pt
  \centering
  \begin{threeparttable}
  \caption{Statistics of the  datasets over the inductive setting.}
   \label{tabled2}
    \begin{tabular}{cccccccc}
    \toprule
    \multirow{2}{*}{ }&
    \multicolumn{4}{c}{\quad WN9-IMG-TXT}&\multicolumn{3}{c}{FB-IMG-TXT}\cr
    \cmidrule(lr){1-8}   &  
    & \#Ent  & \#Rel & \#Triplets & \#Ent  & \#Rel & \#Triplets \cr
    \midrule
    V1 &Training &420&8  &760 &2848 &342 &17561\cr
    V1 &Ind\_test &270 &8 &401  &2002&263 &3325\cr
    \midrule
    V2 &Training&654&9&1465  &3205 &631 &35184  \cr
    V2 &Ind\_test&406 &8  &814  &2111 &343 &6064 \cr
    \midrule
    V3 &Training &658 &9 &2180  &3716  &750 &52544\cr
   V3 &Ind\_test &581 &8  &1442  &2419 &254 &10623 \cr

    \bottomrule
    \end{tabular}
    \end{threeparttable}
 \vspace{-0.2cm}
\end{table}
Following MMKGR \cite{6:MTRL, 7:MMKGR}, we use WN9-IMG-TXT and FB-IMG-TXT to verify the reasoning performance under the transductive setting. 
Each entity in these MKGs contains three modal information: structure, image, and text. Specifically, the relation triplets and textual descriptions  are extracted from WordNet and Freebase. To extract the image features of the entities, 10 images and 100 images are crawled for each entity in WN9-IMG-TXT and FB-IMG-TXT, respectively \cite{6:MTRL}. Statistics are shown in Table ~\ref{tabled1}. 

To perform inductive reasoning in MKGs, we construct new inductive benchmark datasets by extracting disjoint subgraphs from WN9-IMG-TXT and FB-IMG-TXT. In particular, each dataset contains a pair of graphs: \emph{training graph} and \emph{ind\_test graph}. Following \cite{10:GraIL}, to generate the training graph, we first uniformly sample several entities as the root nodes, and then conduct the union of k-hop neighbor triplets around the roots. Next, we set the maximum number of samples at each hop to prevent the exponential growth of new neighbors. Finally,  we remove the training graph from the whole graph and sample the test graph using the same procedure. In fact, the above division method destroys the link distribution in the original graph and reduces the number of triplets. 
Therefore, for a robust evaluation, we adjust the parameters of the above procedure to sample 5\%, 10\%, and 15\% of the original graph (i.e., WN9-IMG-TXT and FB-IMG-TXT) to construct datasets with different sizes \cite{12:LAN}. In summary, the model is trained on the training graph and tested on the ind\_test graph in every version. Note that, (1) the two graphs have disjoint sets of entities, (2) training graphs contain all relations present in ind\_test graphs. 

\subsection{Evaluation Protocol}
To evaluate the reasoning performance of TMR over inductive and transductive settings, we adopt the mean reciprocal rank (MRR) and Hits@N to report experimental results, which are common metrics for MKGR 
 \cite{6:MTRL, 7:MMKGR}.

\begin{table*}
\scriptsize
\setlength\tabcolsep{3pt}
\center
  \caption{
Inductive link prediction results for different versions of MKGs.}
 \vspace{-7pt}
\begin{tabular}
{c|ccccccccc|ccccccccc}
\toprule
\multicolumn{1}{c|}{\multirow{3}{*}{ }} & \multicolumn{9}{c|}{WN9-IMG-TXT}& \multicolumn{9}{c}{FB-IMG-TXT}\\ 
 & \multicolumn{3}{c}{V1}  & \multicolumn{3}{c}{V2} & \multicolumn{3}{c|}{V3}  & \multicolumn{3}{c}{V1}  & \multicolumn{3}{c}{V2}  & \multicolumn{3}{c}{V3}\\ 
  & MRR & Hits@1    & Hits@10 & MRR & Hits@1    & Hits@10& MRR & Hits@1    & Hits@10 & MRR & Hits@1    & Hits@10 & MRR & Hits@1    & Hits@10  & MRR & Hits@1    & Hits@10\\ \midrule
  
MMKGR & 27.0 & 25.1  & 30.2 & 27.2 & 25.5  & 30.6 & 29.3 & 26.9  & 33.3 & 21.4 & 19.3 & 25.3  & 23.3 & 21.7 & 26.5  & 26.0 &23.9  & 29.1\\ 

DRUM& 43.4 & 40.5  & 46.0 & 45.2& 41.6  &48.7 & 48.4 & 45.7  & 51.0 & 35.6 & 32.7  &38.2  & 37.8 & 34.8 & 39.7  &40.1 & 38.6  & 43.2\\ 

CoMPILE& 45.2 & 41.7  & 47.3 & 47.0 & 43.9  & 50.8 & 49.1 & 47.6  & 53.2  & 36.9 & 34.5  & 39.3  &39.8 & 35.9 & 40.9  &42.3 & 39.9  & 44.4\\ 
    
MorsE & 48.2 & 45.8  & 52.4  & 50.3 & 48.6  & 53.1 & 54.2 & 52.1  & 56.7 & 38.3 & 36.6  &41.2  &40.7 &38.3 &43.1  & 44.2 &42.1  & 46.2\\

RED-GNN & {\underline{51.2}} & {\underline{49.3}}  
& {\underline{54.3}} & {\underline{53.3}} & {\underline{50.2}}  &{\underline{56.8}} & {\underline{56.1}} & {\underline{54.8}}  & {\underline{59.2}} & {\underline{40.5}} & {\underline{38.4}} & {\underline{42.4}}  &{\underline{43.3}} &{\underline{41.1}}  & {\underline{45.2}} &{\underline{46.0}} & {\underline{44.2}}  &{\underline{48.4}}\\ 
      
TMR & {\bf64.9} & {\bf62.3}  & {\bf69.1}  & {\bf67.6} & {\bf64.1}  & {\bf72.0} &{\bf71.1} & {\bf69.7}  & {\bf74.8} & {\bf57.0} & {\bf54.7}  &{\bf59.4}  &{\bf60.1} & {\bf57.5}& {\bf62.7}  & {\bf63.3} & {\bf60.8}  & {\bf66.2}\\
\bottomrule
Improv.  &13.7\%     & 13.0\%  & 14.8\% & 14.3\% &13.9\%  & 15.2\% & 15.0\% & 14.9\%  &15.6\%     & 16.5\%  &16.3\% &17\% & 16.8\%  &16.4\% & 17.5\% & 17.3\% & 16.6\% & 17.8\% \\
\bottomrule
\end{tabular}
\end{table*}

\begin{table*}
	\centering
	\caption{Transductive link prediction results on different  MKGs.}
 \vspace{-7pt}
\scriptsize
	\setlength{\tabcolsep}{3.3mm}{
		\begin{tabular}{c|ccc|ccc}
			\toprule
			&  \multicolumn{3}{c|}{WN9-IMG-TXT} & \multicolumn{3}{c}{FB-IMG-TXT} \\ 
			Model         & MRR    & Hits@1 &Hits@10    & MRR        & Hits@1     &Hits@10    \\ \midrule
		
			MMKGR    & 80.2     & 73.6 & 92.8  & 71.3  & 65.8 &82.6 \\
      		TMR-TAIR    & {\underline{83.6}}     & {\underline{76.4}} & {\underline{93.2}}  & {\underline{74.3}}  & {\underline{67.5}} & {\underline{85.4}} \\
   			TMR    & {\bf86.3}     & {\bf79.7} & {\bf 93.7}  & {\bf76.6}  & {\bf71.4} & {\bf87.6} \\
			\bottomrule

		\end{tabular}
	}
	\label{tab:whoiswho}
	\vspace{-3pt}
\end{table*}

\subsection{Baselines and Implementation Details}
To investigate the performance of TMR, two categories of methods are compared. 1) 
Knowledge graph reasoning methods under the induction setting: 
DRUM \cite{13:DRUM}, CoMPILE \cite{11:CoMPILE}, 
MorsE \cite{51:MorsE}, RED-GNN \cite{30:REDGNN}. 2) SOTA MKGR method: MMKGR \cite{7:MMKGR}. Note that, the transductive reasoning methods (i.e., TransE \cite{52:TRANSE} or RLH \cite{50:RLH}) on the traditional KG cannot be applied to MKGR. This is because these methods retrain the embedding from scratch whenever a new entity appears \cite{12:LAN}. In our training stage, some core settings are as follows. The embedding dimension \emph{$d_s$} of relation and history is set to 200, the embedding dimension \emph{$d_i$} of the image feature is set to 128 and 4096 on FB-IMG-TXT and WN9-IMG-TXT respectively, and the embedding dimension \emph{$d_t$} of textual feature is 1000. The maximum reasoning step \emph{L} is 3. The number of additional relation $I$ is 3 and 5 on different versions of WN9-IMG-TXT and FB-IMG-TXT, respectively. The size $N$ is set to 5. We employ a 3-layer GNN to obtain the adaptive topology representation. $\alpha$ in Eq.(29) is set to 0.4 and 0.2 on different versions of WN9-IMG-TXT and FB-IMG-TXT, respectively.

\subsection{Inductive Reasoning on MKGs} 
Reasoning performance over the inductive setting is reported in Table 3 (these scores are in percentage), where the competitive results of the baseline are marked by underlining and the highest performance results are highlighted in bold. Specifically, we have the following insightful analysis based on the experimental results in Table 3. (1) The performance of our proposed TMR outperforms all baselines. This is because TMR not only generates fine-grained entity-independent representations for unseen entities, but also utilizes RARL to eliminate the negative impact of manual rewards and sparse relations on reasoning accuracy. (2) The experimental results of MMKGR are the lowest in the different datasets. This is because MMKGR without induction capability cannot obtain structured representations of unseen entities. Therefore, it is not a reasonable solution to only utilize multi-modal auxiliary information without learning inductive capabilities in the induction setting. (3) The rule-based SOTA model DRUM combines relations to infer unseen  tasks, but this model is limited by the quality of the rules. (4) Similar to DRUM, CoMPILE, MorseE and RED-GNN are designed to infer unseen tasks in traditional knowledge graphs, but they adapt to conduct reasoning without utilizing multi-modal auxiliary features in the inductive setting. This is because the three models learn the inductive ability from the local graph structure information. In particular, the existing SOTA  model RED-GNN remains competitive in all versions of MKGs. This is because RED-GNN recursively encodes multiple relational digraphs with shared edges and preserves the structural patterns at the same time \cite{30:REDGNN}.
In short, the key to performance improvement in the induction setting is to consider both induction capability and the ability to exploit multi-modal data. Note that, TMR is the first model to do this in the domain of MKGR.

\subsection{Transductive Reasoning on MKGs}
In this section, we investigate whether TMR outperforms the SOTA model MMKGR  under the transductive setting. To be consistent with MMKGR, we obtain the pre-trained embeddings of all entities by TransE. This is because unseen entities do not exist under the transductive setting. Specifically, the pre-trained representation $\textbf{e}_i$ of entity $i$ is additionally added into the context features of TMR. Furthermore, a concern is that TMR is fed with pre-trained entity representations do not need entity-independent features generated by the TAIR component that contains topology information in the transductive setting. To eliminate the above concern, we added a variant of TMR-TAIR where the TAIR component is removed (i.e., the context features of TMR do not contain fine-grained entity-independent features).

The experimental results under the transductive setting are shown in Table 4. We have the following observations. (1) The reasoning performance of TMR surpasses that of the SOTA MMKGR on the original MKGs. This demonstrates the flexibility and generalizability of TMR under different settings. (2) The performance of TMR-TAIR is higher than that of MMKGR. This is because the RARL component in TMR-TAIR can eliminate the negative impact of manual rewards and sparse relations on reasoning accuracy. (3) The  performance of TMR-TAIR is lower than that of TMR. A reason is that the entity-independent features generated by the TAIR component aggregate multi-hop topology information. The information can provide reasoning context to improve performance under both inductive and transductive settings.

\subsection{Ablation Study}
The overall goal of the ablation study is to measure the contribution of different components by adding different variants. Figure 3 reports the experimental results of the TMR and its variants. (1) The variant version w/o TAIR only uses multi-modal features of unseen entities where the TAIR component is removed from the TMR. (2) w/o UGAN, in which the unified gate-attention network is removed and basic concatenation operation is added to fuse all features. (3) w/o RARL, a variant version where RARL is removed. To ensure the agent conducts the reasoning, we retain the RL-based reasoner with the  original action space and basic 0/1 reward (i.e., the reward value is set to 1 if the target entity is the ground truth entity. Otherwise, the value is 0).

We have the following observations. (1) TMR has the best performance compared with both variant versions. This validates the effectiveness of all components. (2) The results of w/o TAIR are much lower than that of TMR, which verifies the importance of learning fine-grained entity-independent representation in the inductive setting. (3) Although the input of w/o UGAN includes multi-modal information, its reasoning performance still declines compared with TMR. This is because the concatenation operation cannot generate multi-modal complementary  features. 
(4) After removing the rule-guided action augmentation method and adaptive reward mechanism, the performance of w/o RARA significantly degrades on different datasets. This demonstrates that RARL can eliminate the negative effects of sparse relations and manual rewards on reasoning  performance. 

\begin{figure}
	\centering
	\subfigure[V3-WN9-IMG-TXT]
	{
		\centering
		\includegraphics[width=0.46\linewidth,height=3cm]{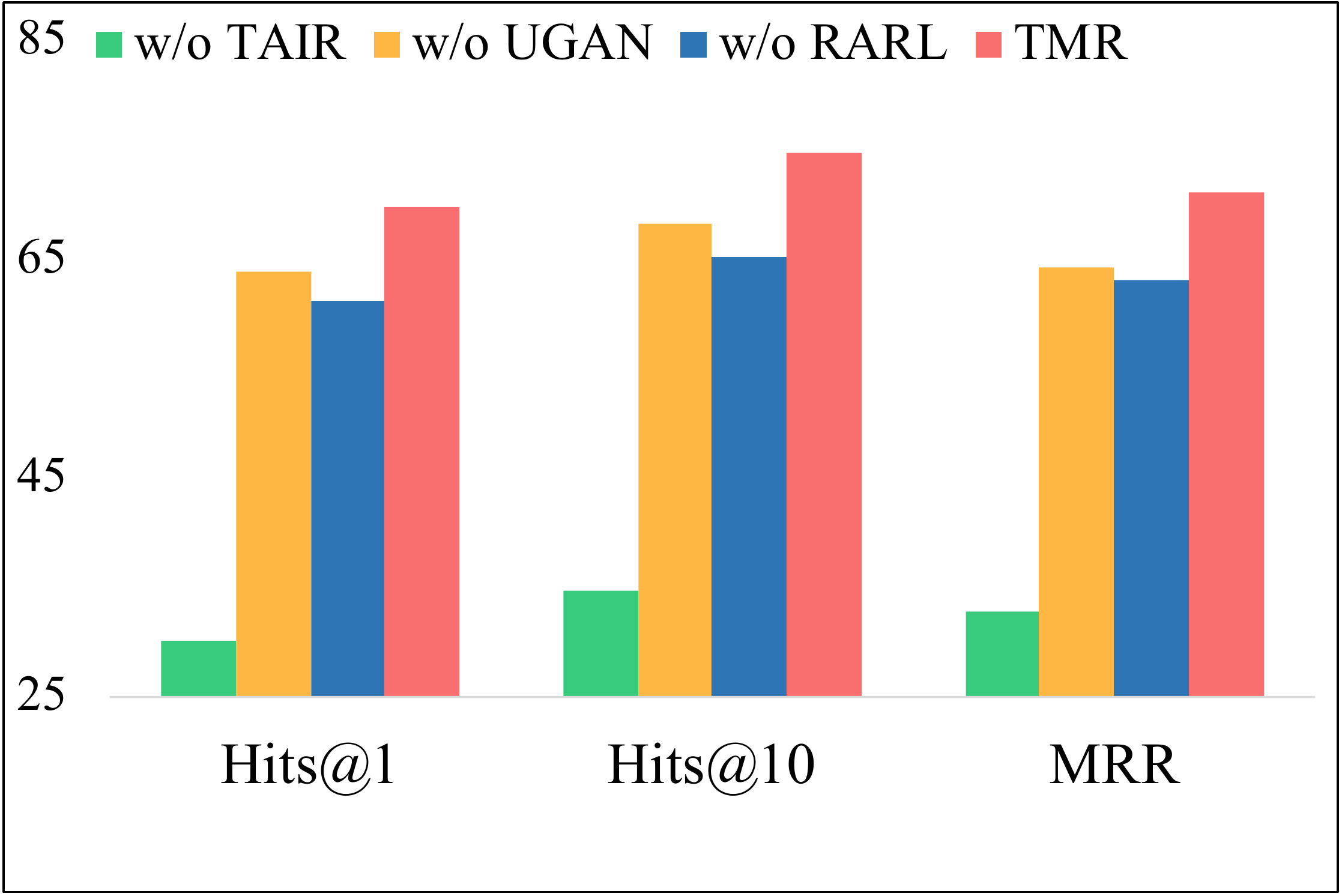}
	}
	\subfigure[V3-FB-IMG-TXT]
	{
		\centering
		\includegraphics[width=0.46\linewidth,height=3cm]{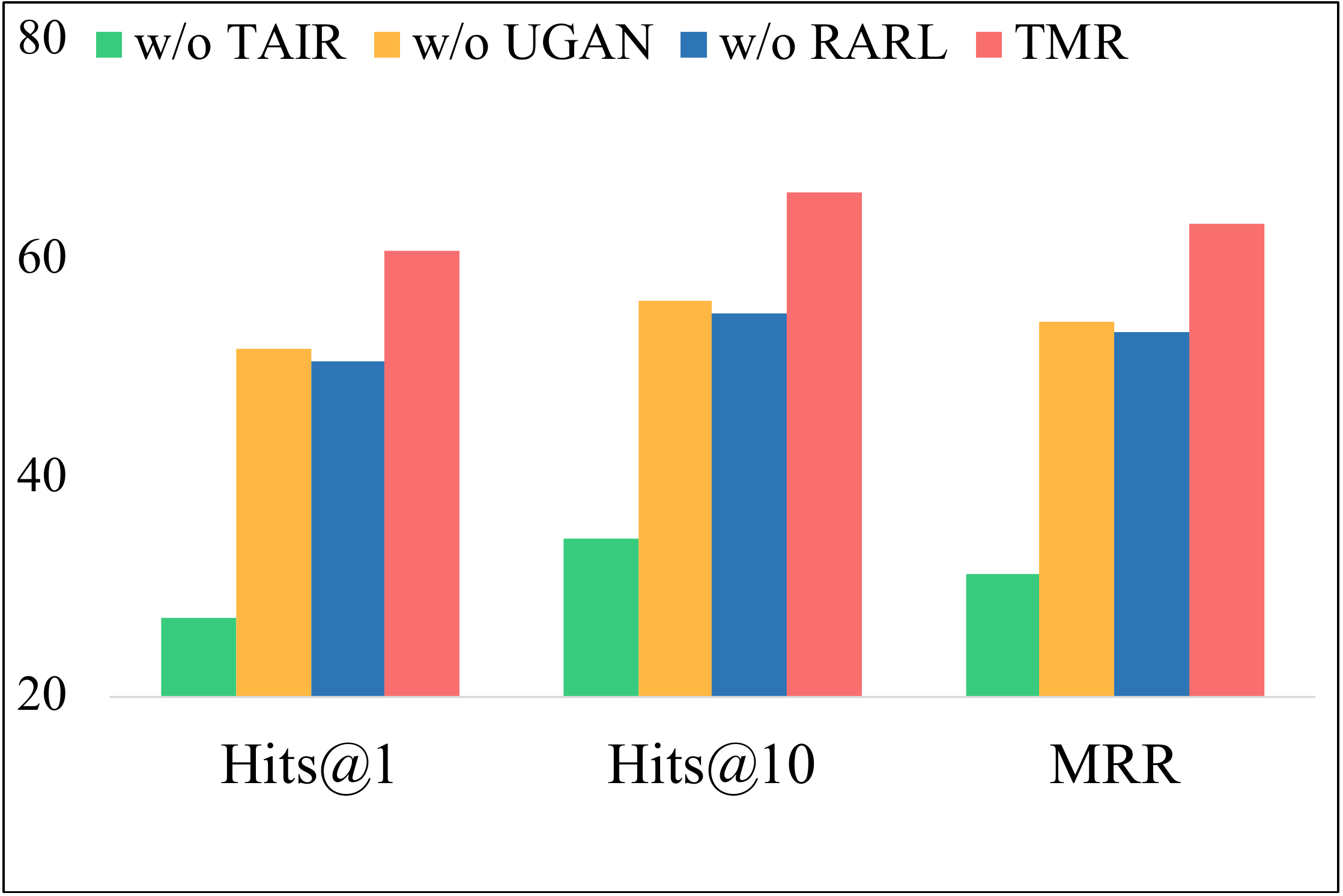}
	}
	\caption{Ablation on different components of the TMR.}
	\label{fig:ablation}
\end{figure}

\begin{figure}
	\centering
	\subfigure[V3-WN9-IMG-TXT]
	{
		\centering
		\includegraphics[width=0.46\linewidth,height=3cm]{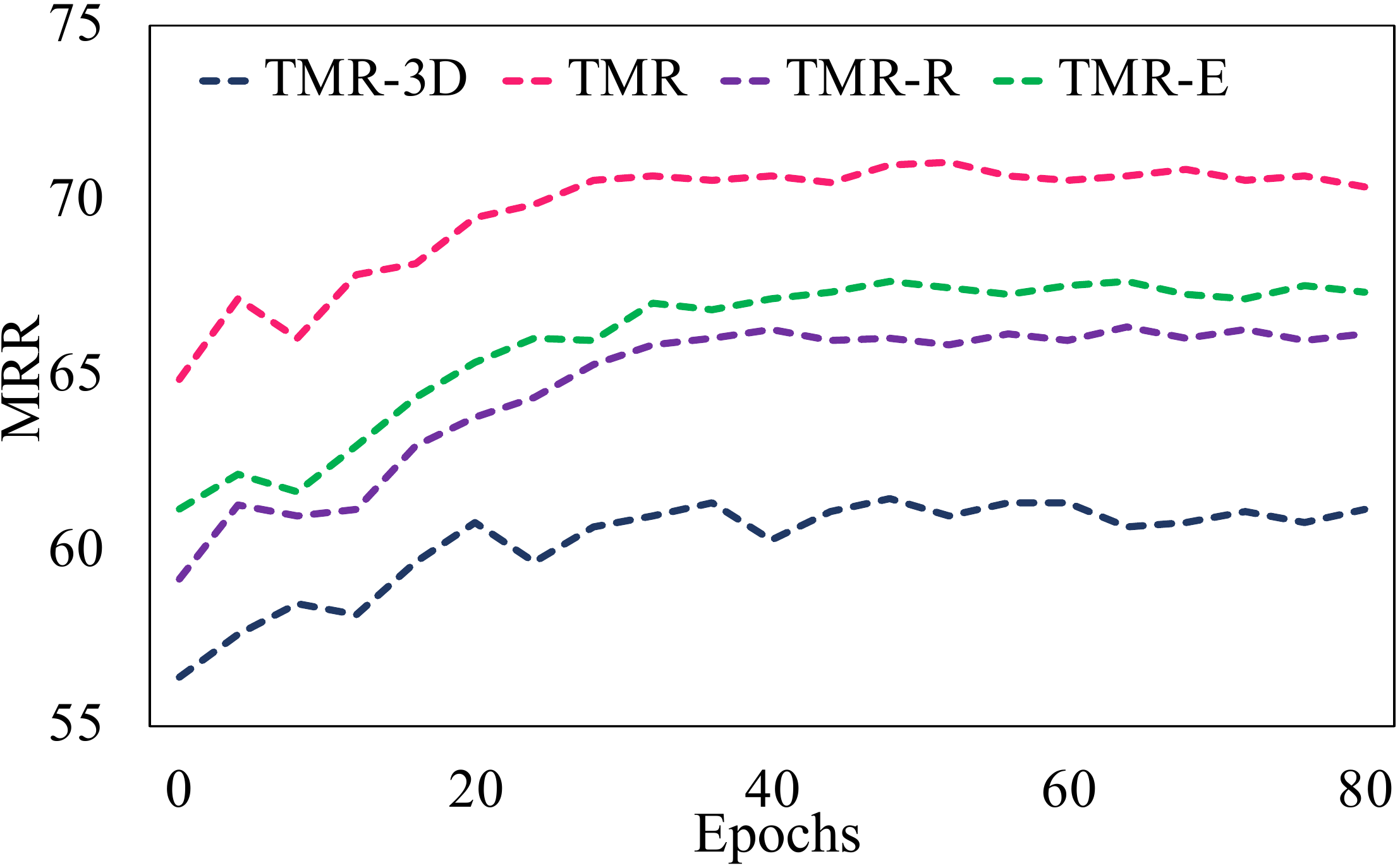}
	}
	\subfigure[V3-FB-IMG-TXT]
	{
		\centering
		\includegraphics[width=0.46\linewidth,height=3cm]{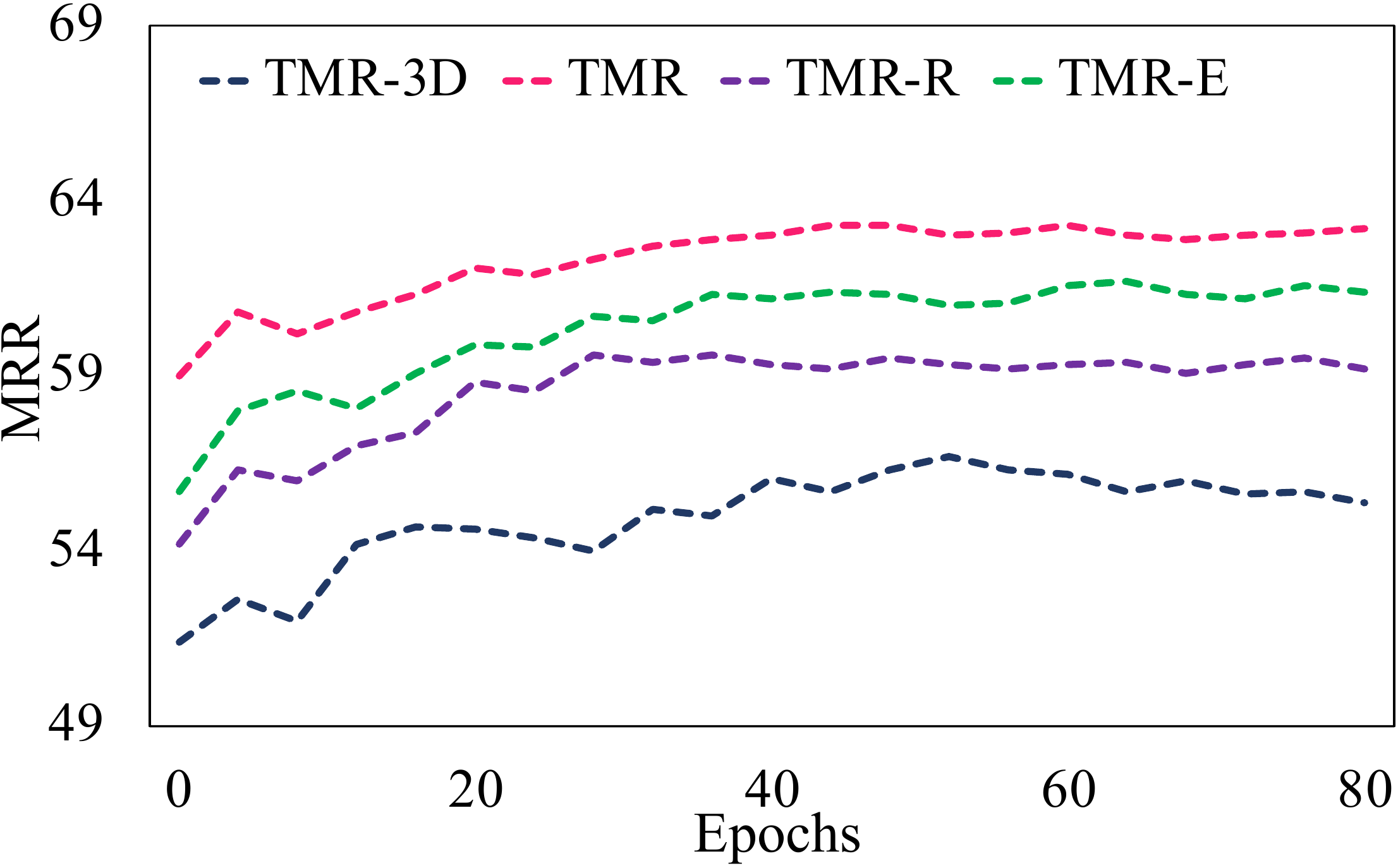}
	}
	\caption{The convergence rate of TMR and the variant model TMR-3D.}
	\label{fig:ablation}
	\vspace{-0.35cm}
\end{figure}

\subsection{Further Analysis}

\subsubsection{Convergence Analysis}
To analyze the convergence rate and reasoning performance between our adaptive reward and manual rewards in MMKGR, we design a variant entitled TMR-3D. Specifically, this variant  removes the \emph{Rule-based Demonstration Sampler} and \emph{Modality-aware Discriminator} in RARL. To ensure that the agent is rewarded, we add the 3D reward mechanism (i.e., the manual reward function in MMKGR). In addition, we design variant models TMR-R and TMR-E by removing the relation-level reward $R_r$ and the entity-level reward $R_e$ in TMR, respectively. This setting can investigate the contribution of different parts of the adaptive reward.  Observed from Figure 4, (1) TMR adopted the adaptive reward has the fastest convergence and the most stable performance on MKGs. This is because the adaptive reward  learned from both path semantic and multi-modal feature automatically eliminate the manual intervention and avoids decision bias on different MKG datasets \cite{54:supred}. (2)  Although TMR-3D can converge slowly, its performance is still unstable. A  reason is that the weak generalizability of manually-designed 3D rewards leads to unstable training on different datasets \cite{24:GAN}. (3) The performance of TMR-R is slightly worse than that of TMR-E, which indicates that $R_r$ has more contribution than $R_e$ in the RARL. 

\begin{figure}
	\centering
	\subfigure[WN9-IMG-TXT]
	{
		\centering
		\includegraphics[width=0.46\linewidth,height=3cm]{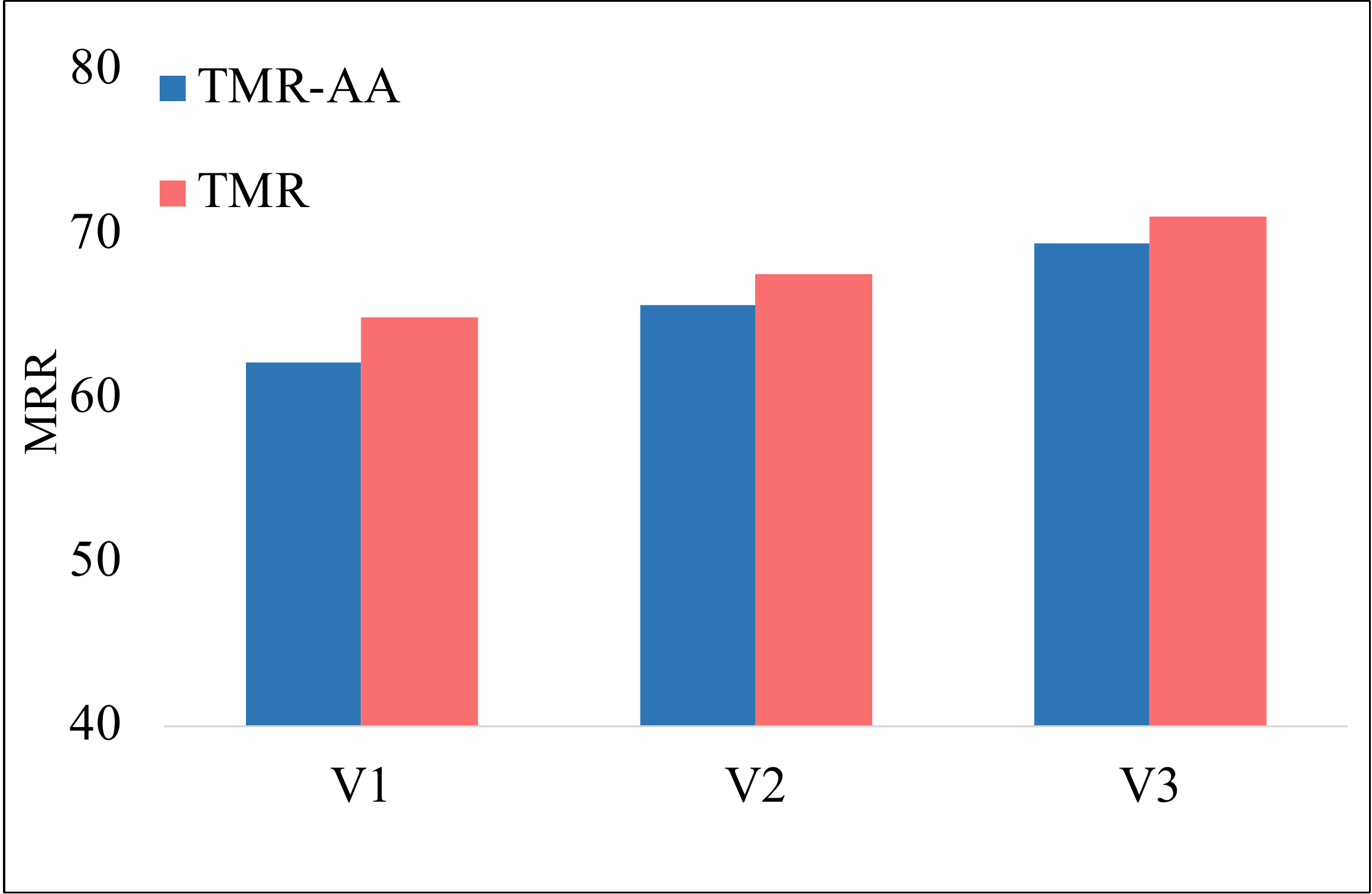}
	}
	\subfigure[FB-IMG-TXT]
	{
		\centering
		\includegraphics[width=0.46\linewidth,height=3cm]{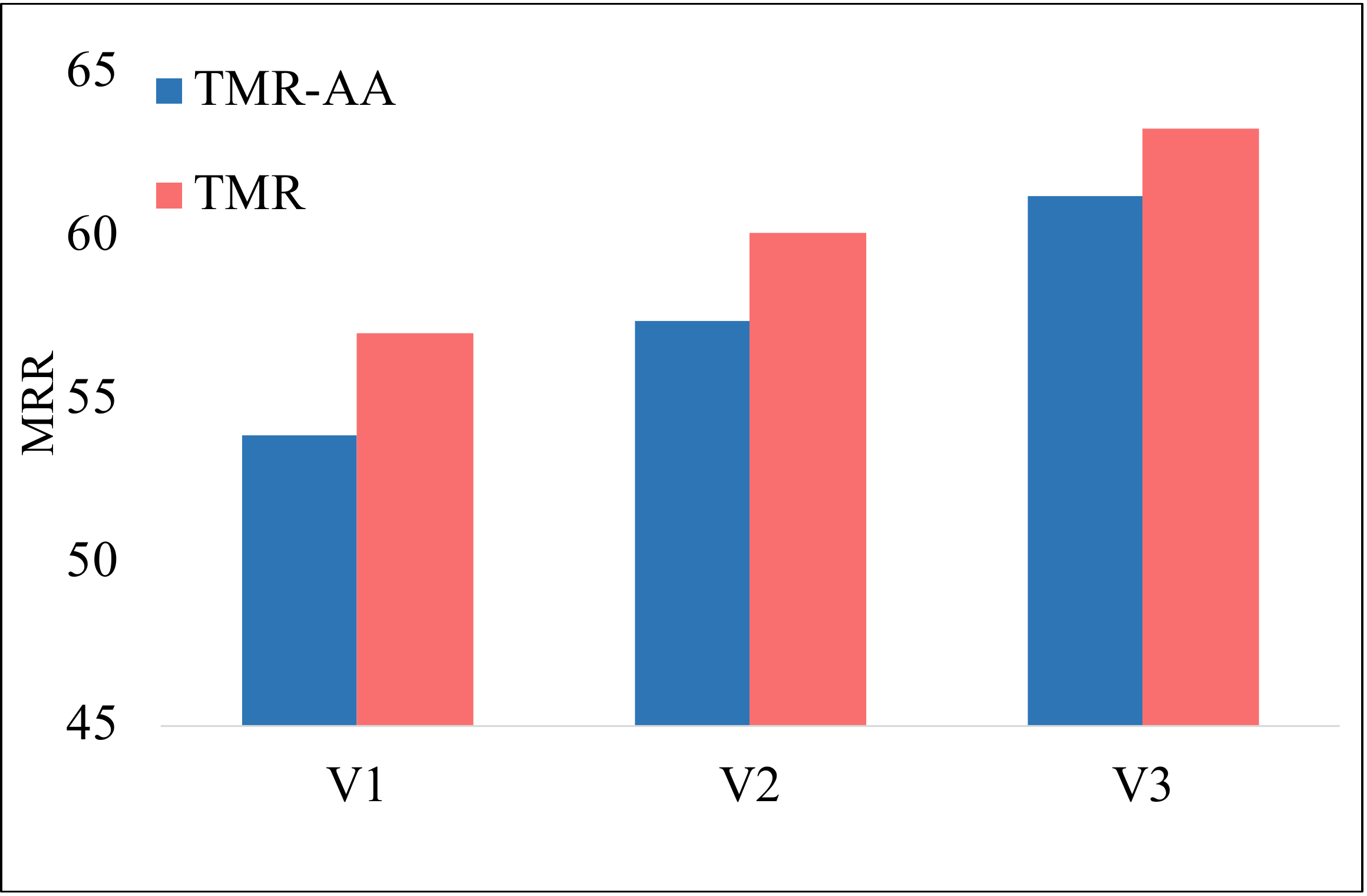}
	}
	\caption{Performance comparison between TMR and TMR-AA.}
	\label{fig:ablation}
	\vspace{-0.2cm}
\end{figure}

\begin{figure}
	\centering
	\subfigure[WN9-IMG-TXT]
	{
		\centering
		\includegraphics[width=0.46\linewidth,height=3cm]{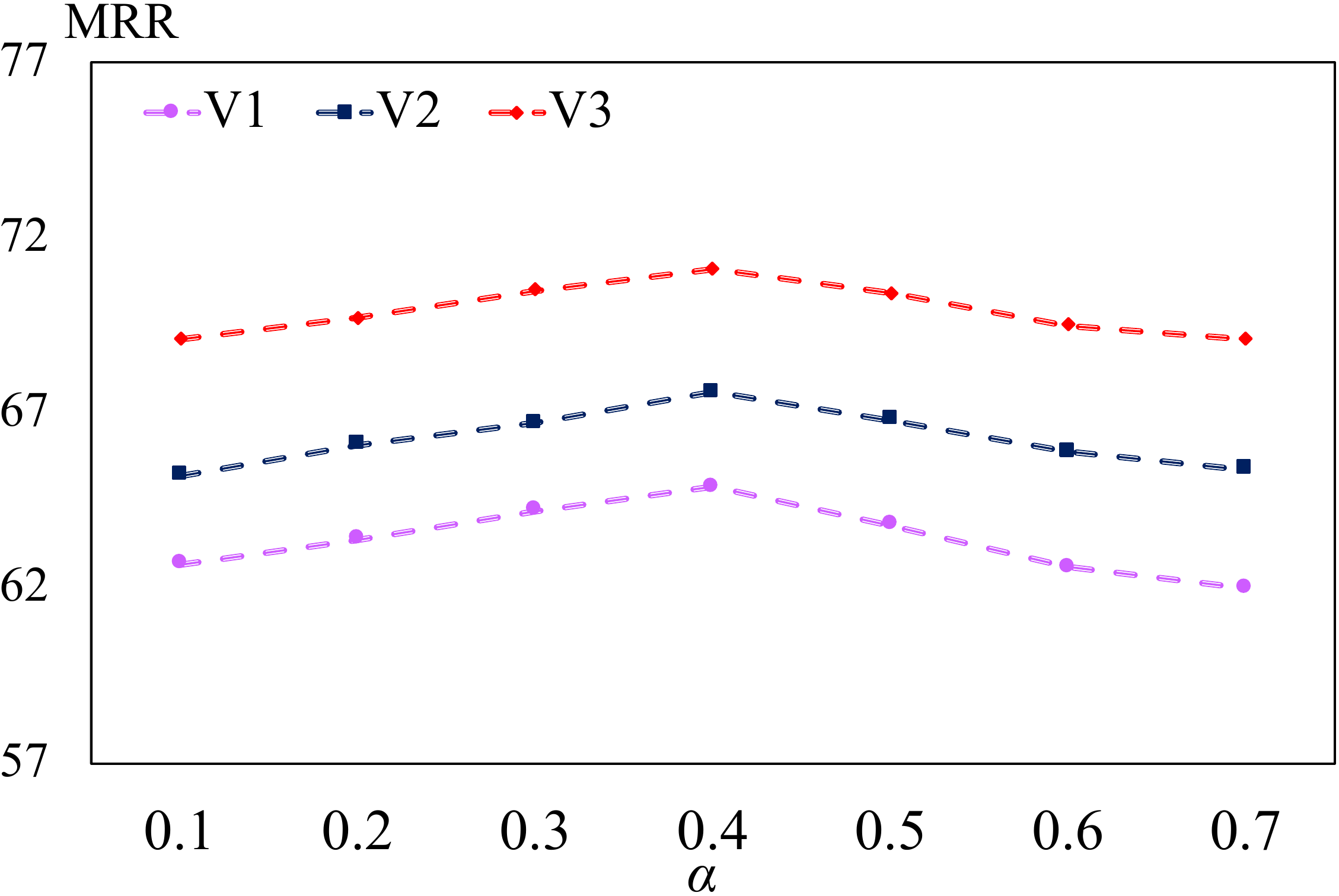}
	}
	\subfigure[FB-IMG-TXT]
	{
		\centering
		\includegraphics[width=0.46\linewidth,height=3cm]{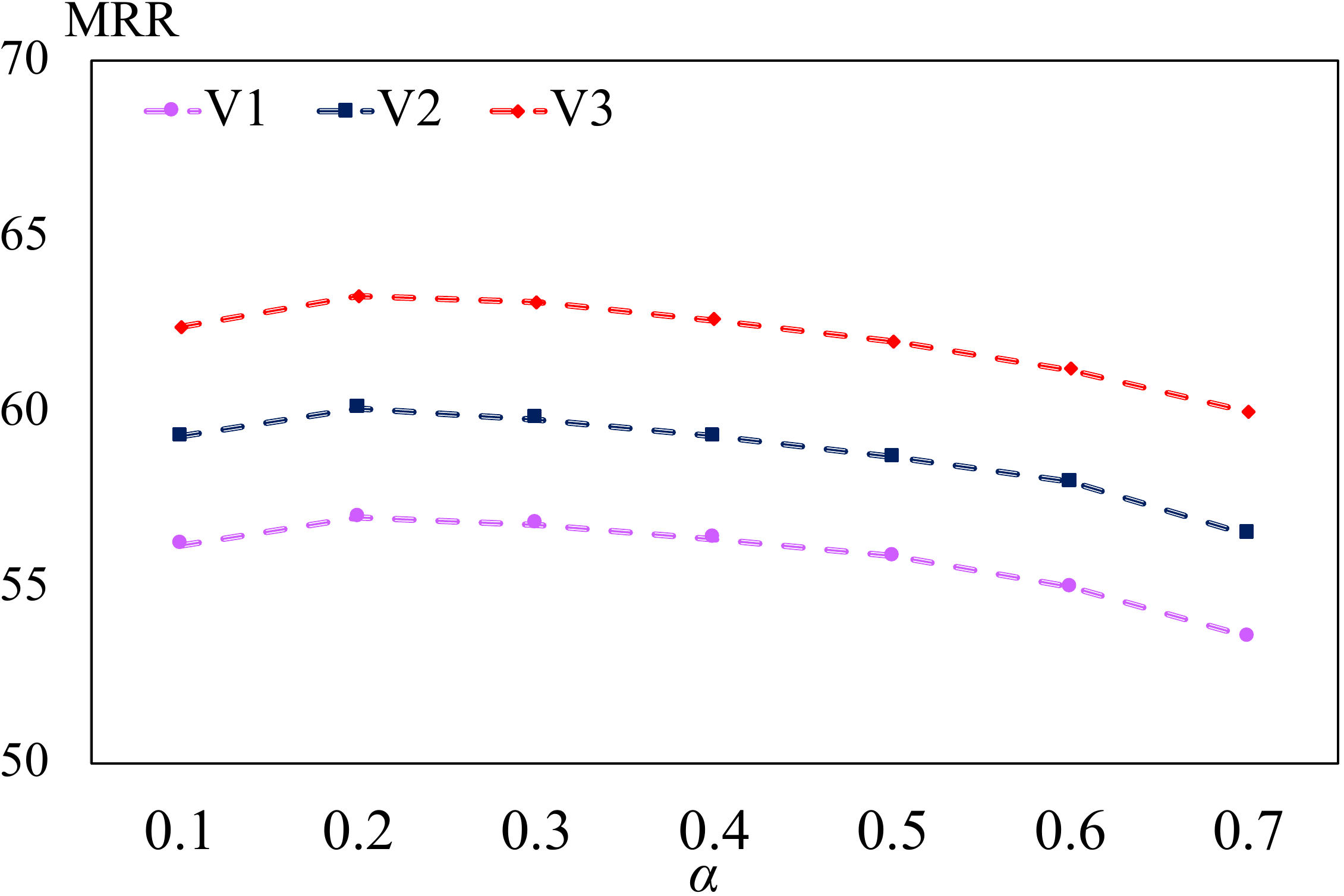}
	}
	\caption{Performance of TMR w.r.t. varied $\alpha$ on different datasets.}
	\label{fig:ablation}
 \vspace{-0.35cm}
\end{figure}

 \subsubsection{Effectiveness Analysis for Action Augmentation}
 To investigate the effectiveness of the action augmentation method  for expanding the latent action space, we design a new variant model TMR-AA by removing the rule-guided action augmentation method in RARL. The experimental results are shown in Figure 5, and we have the following analysis: (1) The performance of TMR has declined to varying degrees in all datasets after removing the action augmentation method, which verifies the effectiveness of the proposed action augmentation method. (2) The performance improvement of the rule-guided action augmentation method on different versions of FB-IMG-TXT is more obvious than that on the different versions of WN9-IMG-TXT. This is because more relations on different versions of FB-IMG-TXT can be used to build more complex reasoning  rules. (3) Compared with adaptive rewards, the performance improvement of the rule-guided action augmentation method is relatively small. One potential reason is that the reasoning processes completely depend on original actions, and the additional actions from this method mainly play an auxiliary role.

\subsection{Key Parameter Analysis}
The balance factor $\alpha$ is an important parameter for our
proposed model TMR. As presented in Figure 6,  0.4 and 0.2  are the optimal values of $\alpha$ on the different versions of WN9-IMG-TXT and FB-IMG-TXT, respectively.  The reward $R_e$ from entity level is assigned small weights, which demonstrates that the semantic correctness of relational paths provides better reasoning clues than multi-modal entities in inductive reasoning tasks.

\section{Related Work}
\subsection{Multi-modal Knowledge Graph}
A traditional KG is essentially a  semantic graph that consists of entities (nodes) and relations (edges). At present, the actual internet data show multi-modal characteristics \cite{40:KGSurvey}. MKGs are developed to incorporate various types of data from different modalities into  KGs \cite{6:MTRL}. A MKG typically includes structural triplets,
and multi-modal data (i.e., texts and images) \cite{56:mmkg}. Common MKGs are IMGpedia \cite{57:Imgpedia}, Richpedia \cite{58:Richpedia}, and FB-Des \cite{44:fusion3}. However, the multi-modal auxiliary data of these MKGs is singular (i.e., image or text). 
 To expand the auxiliary data with one modality,  WN9-IMG-TXT and FB-IMG-TXT simultaneously add a number of textual descriptions and images to each entity,
aiming to further enhance the data diversity of the MKGs \cite{6:MTRL, 63:t2}.

\subsection{Multi-modal Knowledge Graph Reasoning}
Since MKGs inherently contain incomplete knowledge, MKGR technology that can synthesize the original knowledge and infer the missing knowledge is particularly important \cite{6:MTRL, 62:t1}. Some studies employ the attention model or concatenation to fuse multi-modal auxiliary features and then adopt TransE to infer missing elements, such as IKRL \cite{59:IKRL} and TransAE \cite{60:TATREANe}, and MTRL \cite{6:MTRL}. However, these methods lack interpretability and are primarily suitable for single-hop reasoning  containing limited information. To address this issue, MMKGR leverages the symbolic compositionality of the multi-step relational path (choices of actions) to infer the correct entity  \cite{7:MMKGR, 64:t3}. MMKGR has been proven to be a SOTA model in the field of TKGR.  Its multi-hop reasoning process is as intuitive as “going for a walk”, which naturally forms an explainable provenance for MMKGR.

\subsection{Inductive Reasoning on Knowledge Graph}
The inductive setting is receiving increasing attention since unseen entities are emerging in KGs. Therefore, completing knowledge reasoning in an inductive setting is a practical application. Several methods are proposed to solve this problem. Rule-based methods can leverage the logical rules of existing knowledge to infer new facts, because the rules  are independent of specific entities \cite{13:DRUM}. In addition, GraIL \cite{10:GraIL} and CoMPILE \cite{11:CoMPILE} aim to generalize to unseen entities and improve reasoning performance by subgraph extraction, but the enclosing subgraphs cannot learn relational structures so as to weaken the inductive capability.  Inspired by the  powerful graph modeling capabilities of, SOTA models like MorsE \cite{51:MorsE} and RED-GNN \cite{30:REDGNN} utilize GNNs to aggregate topological structures and mine existing neighbor information, which is a promising method for inductive reasoning. However, these methods still have limitations: (1) They do not extract fine-grained entity-independent features related to the query, which restricts their inductive capacity. (2) Lack of ability to utilize multi-modal auxiliary information in MKGR. 

\section{Conclusion}
In this paper, 
we propose TMR as a solution to conduct TKGR in both inductive and transductive settings. Specifically, TMR mainly includes TAIR and RARL. TAIR learns fine-grained entity-independent representation from query-related topology knowledge to represent unseen entities. RARL eliminates the negative impact of sparse relations and artificial rewards on reasoning accuracy by introducing additional actions and adaptive rewards. To ensure that the entities in the training and testing sets are disjoint under the inductive setting, we construct six MKG datasets with varying scales. Experimental results demonstrate the superior performance of our proposed model compared to existing baselines across different settings. 

\bibliographystyle{IEEEtranS}
\bibliography{reference}

\begin{IEEEbiography}[{\includegraphics[width=1in,height=1.25in,clip,keepaspectratio]{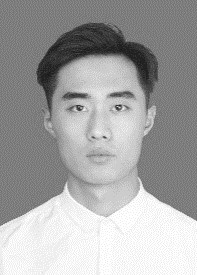}}]{Shangfei Zheng} received the master’s degree in software engineering from Shandong Normal University, China, in 2020. He is currently pursuing the PhD degree with the School of Computer Science \& Technology, Soochow University, China. His main research interests include knowledge graph, multi-modal machine learning and reinforcement learning.
\vspace{-1.65cm}
\end{IEEEbiography}

\begin{IEEEbiography}[{\includegraphics[width=0.9in,height=1.25in,clip,keepaspectratio]{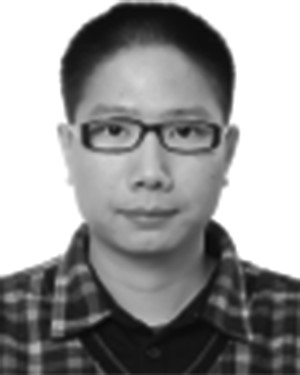}}]{Hongzhi Yin} received the PhD degree in computer science from Peking University, in 2014. He works as an ARC Future Fellow, associate professor, and director of the Responsible Big Data Intelligence Lab (RBDI) at The University of Queensland, Australia. He has made notable contributions to predictive analytics, recommendation systems,  graph learning, and decentralized and edge intelligence. He has published 250+ papers with an H-index of 62.
\vspace{-1.65cm}
\end{IEEEbiography}
\begin{IEEEbiography}[{\includegraphics[width=0.9in,height=1.25in,clip,keepaspectratio]{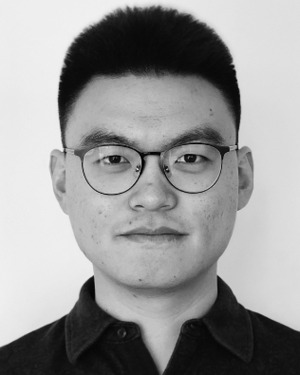}}]{Tong Chen} received the PhD degree in computer science from the University of Queensland, in 2020. He is currently a lecturer in business analytics with the Data Science Group, School of ITEE, University of Queensland. His research interests include data mining, recommender systems, user behavior modelling, and predictive analytics.
\vspace{-1.65cm}
\end{IEEEbiography}
\begin{IEEEbiography}[{\includegraphics[width=0.9in,height=1.25in,clip,keepaspectratio]{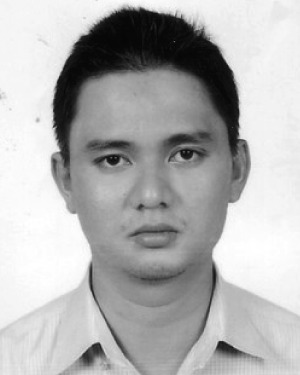}}]{Quoc Viet Hung Nguyen} received the Ph.D. degree from EPFL, Switzerland. He is currently a Senior Lecturer with Griffith University, Australia. He has published several articles in top-tier venues, such as SIGMOD, VLDB, SIGIR, KDD, AAAI, ICDE, IJCAI, JVLDB, TKDE, TOIS, and TIST. His research interests include data integration, data quality, information retrieval, trust management, recommender systems, machine learning, and big data visualization.
\vspace{-1.65cm}
\end{IEEEbiography}
\begin{IEEEbiography}[{\includegraphics[width=1in,height=1.25in,clip,keepaspectratio]{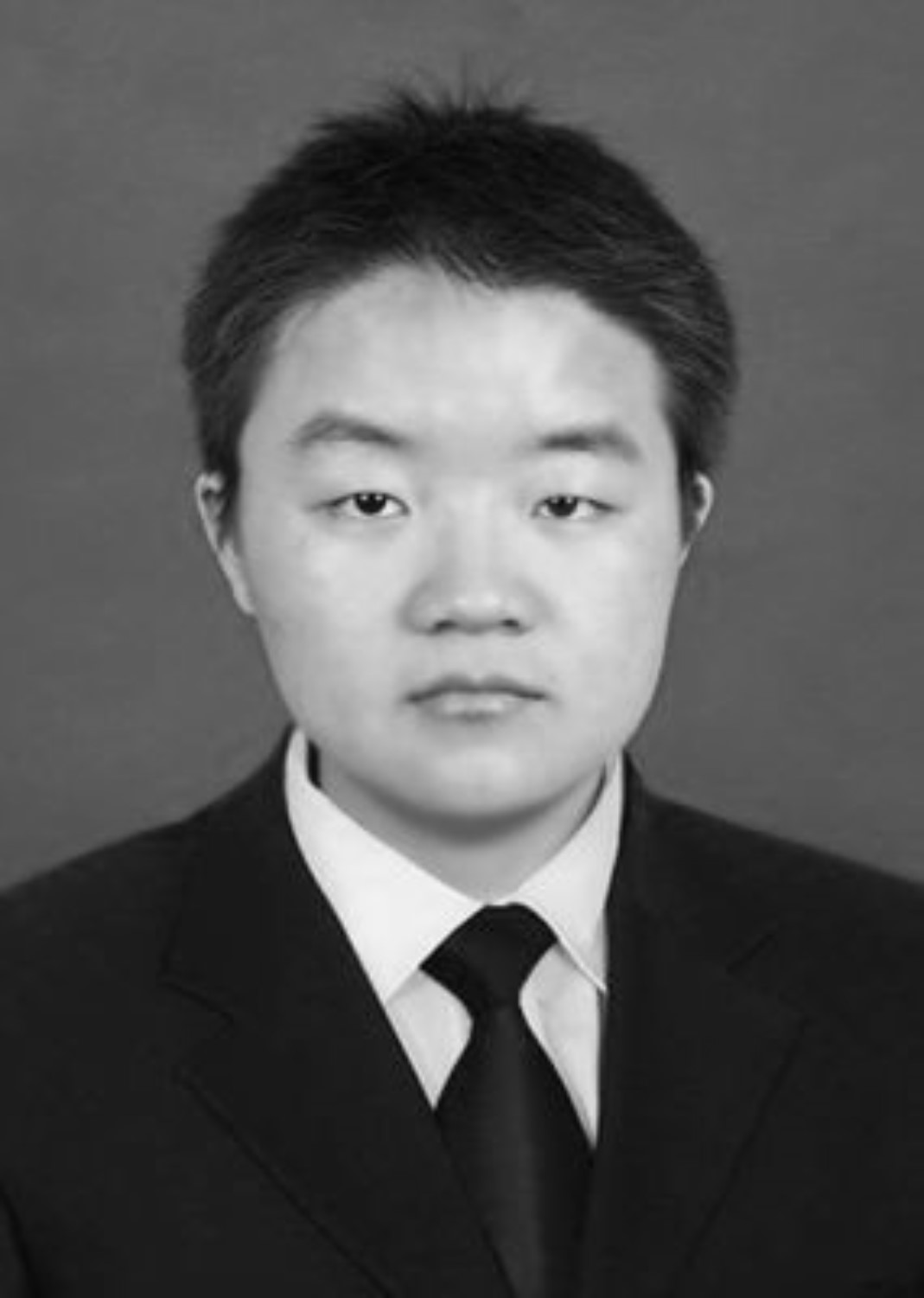}}]{Wei Chen} received the PhD degree in computer science from Soochow University, China, in 2018. He is currently an associate professor with the School of Computer Science and Technology, Soochow University, China. His research interests include spatiotemporal data analysis, heterogeneous network analysis, knowledge graph.
\vspace{-1.65cm}
\end{IEEEbiography}
\begin{IEEEbiography}[{\includegraphics[width=0.9in,height=1.25in,clip,keepaspectratio]{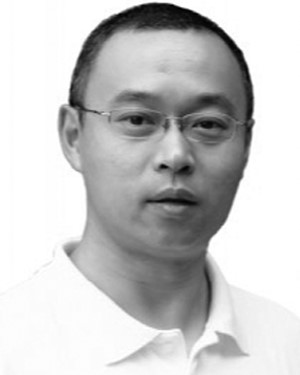}}]{Lei Zhao} received the PhD degree in computer science from Soochow University, China, in 2006. He is currently a professor with the School of Computer Science and Technology, Soochow University, China. His research interests include graph databases, social media analysis, query outsourcing, parallel, and distributed computing.
\vspace{-1.5cm}
\end{IEEEbiography}



%







\end{document}